\documentclass{article}

\usepackage[preprint]{neurips_2026}
\usepackage{comment}

\usepackage{longtable}
\usepackage{booktabs}  
\usepackage{array, makecell}
\usepackage{xcolor}    
\usepackage{pifont}    
\usepackage[utf8]{inputenc} 
\usepackage[T1]{fontenc}    
\usepackage{hyperref}       
\usepackage{url}            
\usepackage{booktabs}       
\usepackage{amsfonts}       
\usepackage{nicefrac}       
\usepackage{microtype}      
\usepackage{xcolor}         
\usepackage{amsmath}
\usepackage{graphicx}
\usepackage{makecell}      
\usepackage[table]{xcolor}  
\definecolor{lightgrayline}{gray}{0.85}  
\usepackage{subcaption}
\usepackage{tcolorbox}
\usepackage{pifont}
\usepackage{multirow}
\usepackage{amssymb}
\usepackage{array} 
\usepackage{enumitem}
\usepackage{makecell}
\usepackage{siunitx}
\tcbuselibrary{breakable, skins}
\usepackage{caption}   
\usepackage{algorithm}
\usepackage{algpseudocode}
\usepackage{tablefootnote}

\newcommand{\bcirc}[1]{%
  \tikz[baseline=(char.base)]{
    \node[shape=circle, fill=black, inner sep=1.2pt] (char)
      {\color{white}\scriptsize\bfseries #1};
  }%
}

\newif\ifshowcomments
\showcommentstrue 
\ifshowcomments
    \newcommand{\jz}[1]{{\color{blue}[JZ: #1]}}
    \newcommand{\ry}[1]{{\color{orange}[RY: #1]}}
    \newcommand{\sh}[1]{{\color{purple}[SH: #1]}}
\else
    \newcommand{\jz}[1]{}
    \newcommand{\ry}[1]{}
    \newcommand{\sh}[1]{}
\fi

\newcommand{\proposed}{AutoSaddler}

\title{\texttt{\proposed}: Automatic Harness Optimization with Durable Updates from Agent Execution Traces}

\author{
  Sungho Park\textsuperscript{1}\thanks{Work done during an internship at Microsoft.} \quad
  Wonjoong Kim\textsuperscript{2}\footnotemark[1] \quad
  Rongyuan Tan\textsuperscript{3}\footnotemark[1] \quad
  Jue Zhang\textsuperscript{4}\thanks{Corresponding authors.} \quad
  Wook-Shin Han\textsuperscript{1}\footnotemark[2] \\
  \textbf{Pengfei Gao}\textsuperscript{4} \quad
  \textbf{Chanyoung Park}\textsuperscript{2} \quad
  \textbf{Yongqiang Yao}\textsuperscript{4} \quad
  \textbf{Rao Fu}\textsuperscript{4} \quad
  \textbf{Elsie Nallipogu}\textsuperscript{4} \\
  \textbf{Qingwei Lin}\textsuperscript{4} \quad
  \textbf{Saravan Rajmohan}\textsuperscript{4} \quad
  \textbf{Dongmei Zhang}\textsuperscript{4} \\[0.5em]
  \textsuperscript{1}POSTECH \quad
  \textsuperscript{2}KAIST \quad
  \textsuperscript{3}Southern University of Science and Technology \quad
  \textsuperscript{4}Microsoft \\
  \texttt{juezhang@microsoft.com} \quad
  \texttt{wshan@dblab.postech.ac.kr}
}

\begin{document}

\maketitle

\begin{abstract}
LLM agents remain unreliable on long-horizon tasks, where small local failures can compound over extended interactions and lead to overall task failure. Although external harnesses can substantially improve robustness, harness design remains a manual and expensive process that requires searching over a large space of prompts, tool configurations, and control logic. We propose \textbf{\proposed}, an automatic harness optimization framework that formulates harness improvement as an offline learning problem and iteratively updates the harness using failure signals from mini-batches. \proposed{} combines failure-trace diagnosis, structured patch generation that treats the harness as code, and validation-based update selection. Experiments on GAIA2, SWE-Bench Pro, and Terminal-Bench 2.0 show that \proposed{} substantially improves agent performance over the corresponding base harnesses, achieving gains of \textbf{9.0, 9.6, and 10.0} percentage points, respectively. Ablation studies further suggest that effective harness optimization benefits from three ingredients: deep debugging rather than shallow reflection, targeted modifications rather than unconstrained editing, and generalization-aware selection rather than trajectory-specific repair. Together, these results suggest that automatic harness optimization is a promising path toward more performant and reliable agent systems. Project website and code will be available at~\href{this URL}{https://aka.ms/AutoSaddler-website}.

\end{abstract}
\begin{figure}[!h]
\centering
    \begin{subfigure}[b]{0.49\linewidth}
        \centering
        \includegraphics[width=\linewidth]{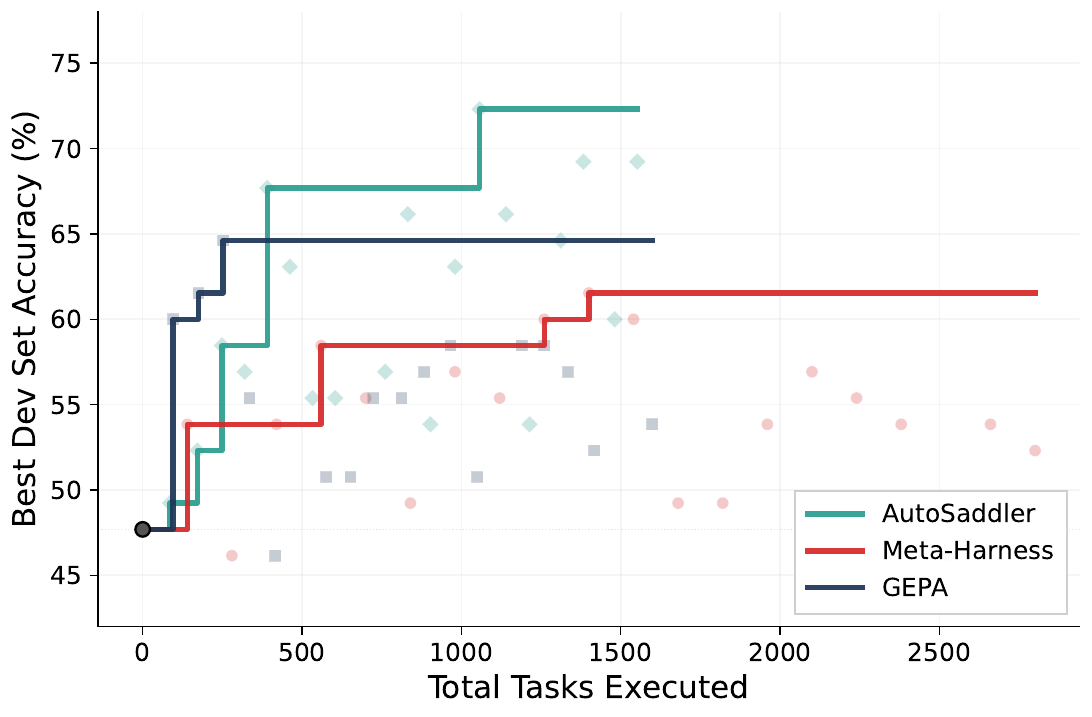}
        \caption{Compute Efficiency}
        \label{fig:compute_efficiency}
    \end{subfigure}
    \hfill
    \begin{subfigure}[b]{0.49\linewidth}
        \centering
        \includegraphics[width=\linewidth]{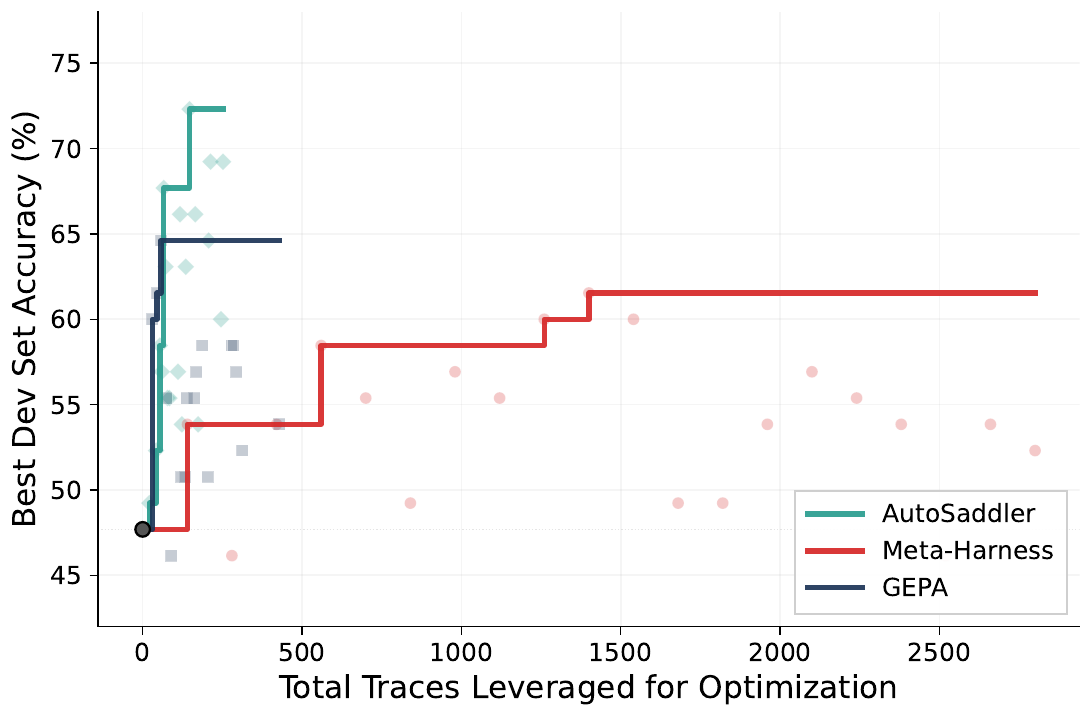}
        \caption{Learning Efficiency}
        \label{fig:learning_efficiency}
    \end{subfigure}
    \caption{\textbf{Comparison of optimization performance and efficiency on GAIA2.} \textbf{(a)} \proposed{} reaches \textbf{72.3\%} dev accuracy with ${\sim}$\textbf{1,000} total task executions, whereas GEPA and Meta-Harness saturate at \textbf{64.6\%} and \textbf{61.5\%}, respectively, despite consuming ${\sim}$\textbf{2{,}800} task executions. \textbf{(b)} When measured by the number of execution rollouts leveraged for optimization, \proposed{} achieves its best performance after consuming only 147 traces, ${\sim}$\textbf{10}$\times$ fewer than Meta-Harness (1,400 traces).
    }

    \label{fig:optimization_comparison}
\end{figure}

\section{Introduction}
\label{sec:introduction}

Large Language Models (LLMs) have improved rapidly in capability in the past years. Despite this progress, they continue to exhibit ``jagged intelligence'': strong performance on some tasks coexists with poor performance on others that appear closely related or even simpler. This unevenness creates a fundamental reliability challenge for agentic applications, especially in autonomous, multi-step, and long-horizon settings, where success depends on sustained competence across many consecutive decisions. In response, an emerging line of work seeks to build external harness layers around LLMs to make agent behavior more robust on long-horizon tasks~\cite{zhou2026externalization, anthropic-harness-practice-1, anthropic-harness-practice-2, openai-harness-practice, langchain-harness-practice}. Empirical evidence suggests that a well-designed harness can substantially improve agent performance~\cite{langchain-harness-practice}.

Manual harness tuning, however, is time-consuming and difficult to scale. It requires exploring a large design space, including prompt specifications, tool configurations, and other system-level choices. Moreover, evaluating each candidate harness is costly, since the agent may need to execute many steps before task success or failure becomes clear. Analyzing the resulting long-horizon trajectories is itself nontrivial and often requires substantial manual effort. These challenges create a strong need for automated harness optimization methods that can efficiently adapt the harness when switching to a new underlying LLM or deploying the agent in a new domain.

In this work, we formulate automatic harness optimization as an offline learning problem and introduce \textbf{\proposed}, a framework that iteratively refines the harness using failure signals from batches of training tasks. At each iteration, we first construct a candidate harness from the accumulated exploration history and evaluate it on the training examples in the current mini-batch. We then diagnose failed agent trajectories and use the resulting insights to guide patch generation, treating the harness itself as code. For each accepted harness update, we further assess generalization on a validation set. Together, these components enable \proposed~to optimize harnesses in a principled and scalable manner, while promoting updates that generalize beyond the observed training samples.

We evaluate \proposed{} on three agent benchmarks: GAIA2~\cite{GAIA2}, SWE-Bench Pro~\cite{SWE-Bench-Pro}, and Terminal-Bench 2.0~\cite{TerminalBench2}. Across all three benchmarks, \proposed{} substantially improves long-horizon agent performance. It surpasses the corresponding base harness by  \textbf{9.0} percentage points on GAIA2, \textbf{9.6} points on SWE-Bench Pro, and \textbf{10.0} points on Terminal-Bench 2.0, and exceeds the strongest automated baseline by \textbf{7.4}, \textbf{4.4}, and \textbf{6.7} points, respectively. Ablations and qualitative analysis further support the design rationale of \proposed{}, showing that effective harness optimization depends on three key ingredients. First, \textbf{in-depth diagnosis} is necessary because long-horizon failures require deep debugging rather than shallow reflection on agent execution failures. Second, \textbf{structured intervention} is important because the large and diverse harness space favors targeted modifications over unconstrained editing. Third, \textbf{generalization-aware selection} is critical because improving performance over a task distribution requires retaining broadly useful updates rather than repairing a single trajectory.

Our main contributions are summarized as follows:
\begin{itemize}[leftmargin=1em,itemsep=0.25em,topsep=0.25em]
    \item We formulate automatic harness optimization as an offline learning problem over prompts, tools, and middleware, leveraging failure signals from agent execution traces.

    \item We introduce \proposed, which combines evidence-grounded diagnosis, structured patching, and generalization-aware selection to produce durable harness updates.

    \item We conduct extensive experiments on three challenging agent benchmarks, demonstrating the effectiveness of the proposed approach in improving long-horizon agent performance.
\end{itemize}
\section{Related Work}

\paragraph{Auto Prompt Optimization.}
Automatic optimization of LLM-based systems has been extensively studied in the context of prompt optimization, including gradient-free search~\cite{DBLP:conf/iclr/ZhouMHPPCB23}, textual-gradient and mini-batch methods~\cite{DBLP:conf/emnlp/PryzantI0L0023}, history-aware and evolutionary optimization~\cite{DBLP:conf/iclr/Yang0LLLZC24,DBLP:conf/iclr/Guo0GLS0L0Y24,agrawal2026gepa}, planning-based search~\cite{DBLP:conf/iclr/WangLW0LZJXH24}, programming abstractions~\cite{khattab2024dspy}, autodiff-style optimization~\cite{DBLP:journals/corr/abs-2406-07496}, and Bayesian multi-prompt tuning~\cite{DBLP:conf/emnlp/Opsahl-OngRPBPZ24}. While AutoSaddler draws on this line of work, harness optimization is broader than prompt optimization: long-horizon traces often require deeper diagnosis than simple reflection, and the search space spans prompts, tools, and runtime control logic.

\paragraph{Self-Evolving Agents and Experience-Based Improvement.}
Learning from failure traces is related to experience-based agent improvement~\cite{DBLP:conf/aaai/Zhao0XLLH24} and self-evolving agents~\cite{gao2026a}, where experience is used to build libraries~\cite{DBLP:journals/corr/abs-2511-06449}, create tools~\cite{DBLP:journals/corr/abs-2505-20286,DBLP:journals/corr/abs-2511-13646}, and construct memory, skills, or knowledge bases~\cite{ouyang2026reasoningbank,DBLP:journals/corr/abs-2511-20297,nousresearch_hermes_agent_website}. Unlike these typically online continual-learning approaches, we focus on offline harness optimization for generalization across environments.

Some prior work further emphasizes self-referentiality, i.e., whether the meta-agent coincides with the task agent, as in Darwin-G\"odel Machines~\cite{zhang2026darwin, DBLP:journals/corr/abs-2603-19461}, G\"odel agents~\cite{DBLP:conf/acl/YinWPL0W25}, and Huxley-G\"odel Machines~\cite{wang2026huxleygodel}. In contrast, we do not impose such constraints, as our objective is to improve the
task agent's external harness under a finite rollout budget rather than to model
self-referential dynamics.

\paragraph{Agent Systems and Harness Optimization.} Automatic optimization of the external layers of LLM-based systems builds on a broad line of prior work, including LLM inference hyperparameter tuning~\cite{DBLP:conf/automl/00010A23, DBLP:conf/emnlp/FuQYWZLC0RZ24}, workflow optimization~\cite{DBLP:conf/iclr/ZhangXYTCCZCHWZ25, DBLP:conf/emnlp/ZhuJMYWGBZLRZ25, DBLP:conf/emnlp/Wang0FM25}, and system optimization by treating the system itself as code~\cite{DBLP:conf/icml/ZhangZLS0KW24, DBLP:conf/iclr/HuLC25}. More recently, this line of work has expanded to optimizing the full agent harness~\cite{lee2026meta, lou2026autoharness, pan2026natural, ursekar2026vero}. One prominent work is Meta-Harness~\cite{lee2026meta}, which likewise employs a coding agent to build an end-to-end optimization pipeline. In contrast, we place greater emphasis on structured patching and on a systematic training pipeline that explicitly accounts for generalization.


Recent contemporaneous studies~\footnote{These works appeared during the final stage of this work preparation or the subsequent peer-review period.}
have further expanded automatic harness optimization through observability-driven evolution~\cite{lin2026agentic}, diagnosis- and weakness-guided repair~\cite{chen2026failed,zhang2026self}, Bayesian configuration search~\cite{sengupta2026harbor}, historical-experience recalibration~\cite{guo2026drevo}, self-supervised retrospective optimization~\cite{pan2026evolving}, generalization-oriented constrained evolution~\cite{zhang2026harnesscompass}, experience-driven test-time adaptation~\cite{huang2026memoharness}, and joint optimization of harnesses and model weights~\cite{hebbar2026sia,chen2026co}. Collectively, these works explore complementary aspects of harness optimization, while AutoSaddler emphasizes the joint roles of in-depth diagnosis, structured intervention, and generalization-aware selection.

\paragraph{Agent Trace Failure Diagnosis and Repair.}
Long-horizon agent traces with complex tool interactions have motivated work on failure diagnosis~\cite{DBLP:conf/icml/ZhangY0LHZL0W0W25, DBLP:journals/corr/abs-2509-03312, ma2025automatic, ge2025introducing,zhang2025graphtracer}. These methods primarily target single-trace diagnosis and often assume traces fit within the LLM context window, limiting their applicability to very long traces and mini-batch diagnosis. We instead build on the Claude Agent SDK (CA-SDK)~\cite{CCSDK}, leveraging its file access and context management capabilities. Recent work also studies trace-level interventions for repairing failed runs~\cite{ma2026dover, zhu2025llm}. Unlike such task-specific ``hot fixes'', our goal is to produce persistent harness repairs that improve future behavior.

\paragraph{LLM-Driven Evolutionary Search.}
Our candidate selection method uses LLMs to select and recombine patches from prior candidates based on exploration history, aligning with recent LLM-augmented evolutionary algorithms~\cite{agrawal2026gepa,DBLP:journals/corr/abs-2506-13131,openevolve,yan2026pacevolve,liu2026evox}. We instantiate this idea with CA-SDK-based candidate selection, leaving specialized LLM-driven evolutionary algorithms for future work.

\section{Preliminaries}
\label{sec:preliminaries}

\paragraph{Agentic Task and Execution Trace.}
We consider agentic tasks that require multi-step reasoning, repeated tool use, and environment interaction. Let $x \in \mathcal{X}$ be a task description sampled from a task distribution $\mathcal{T}$, and assume that task instances in $\mathcal{X}$ are stateless and independent.

Conditioned on a harness parameter \(\theta\), executing the task agent on a task
\(x\) induces a stochastic execution process. We write
$
(\tau,\hat y) \sim P_\theta(\cdot \mid x),
$
where \(\tau\) is the execution trace and \(\hat y\) is the final output. Equivalently,
the harness \(H_\theta\) defines the conditional execution distribution
\(P_\theta(\tau,\hat y \mid x)\). This stochasticity captures randomness from LLM
sampling, tool-use decisions, and environment interactions.

\paragraph{Optimization Space: Agent Harness.}
Although the term \textit{harness} is often used colloquially to refer to everything outside the LLM that contributes to effective agent behavior, its precise scope is still evolving in the community~\cite{zhou2026externalization}. In this work, following harness engineering practices in LangChain~\cite{langchain-harness-practice}, we focus on three classes of harness parameters and do not consider other components, such as memory or skill curation, since our setting assumes tasks are largely stateless and independent. Formally, we define the optimization space as:
\begin{equation}
    \theta = (\theta_{\text{prompt}}, \theta_{\text{tool}}, \theta_{\text{middleware}}) \in \Theta , 
\end{equation} \label{eq:optimization_space}
where $\theta_{\text{prompt}}$ specifies the instructions and system prompts, $\theta_{\text{tool}}$ defines the available tools and their interfaces, and $\theta_{\text{middleware}}$ captures runtime control logic, including hooks and agent-loop behavior.

\paragraph{Objective: Budget-Constrained Optimization.}
Our population objective is to maximize expected task performance over the target
task distribution. Let \(\mu(\hat y,y^\ast)\) be a task-level metric, such as
accuracy or pass/fail success, that evaluates the final output \(\hat y\) against
the gold answer \(y^\ast\). 
Since executing a harnessed LLM agent may be stochastic, we define
\[
J(\theta)
=
\mathbb{E}_{(x,y^\ast)\sim\mathcal{T}}
\mathbb{E}_{(\tau,\hat y)\sim P_\theta(\cdot \mid x)}
[\mu(\hat y,y^\ast)],
\qquad
\theta^\ast = \arg\max_{\theta\in\Theta} J(\theta),
\]
where $\theta^\ast$ denotes the optimized harness.

In practice, \(J(\theta)\) is not directly observable and each execution consumes a
rollout. Given a rollout budget \(K\), \proposed~searches over a set of
candidate harnesses \(\mathcal{V}_K \subset \Theta\) explored within that budget.
For a finite dataset \(D\), we use the empirical estimate
\[
\widehat{J}_{D}(\theta)
=
\frac{1}{|D|}
\sum_{(x_i,y_i^\ast)\in D}
\frac{1}{R_i}
\sum_{r=1}^{R_i}
\mu(\hat y_{i,r},y_i^\ast),
\qquad
(\tau_{i,r},\hat y_{i,r})\sim P_\theta(\cdot\mid x_i),
\]
where \(R_i\) is the number of repeated executions for task \(i\). 
\proposed~returns the candidate with the highest
\textit{development-set} empirical score among candidates evaluated within the rollout budget:
\[
\hat{\theta}_{\mathrm{AS}}
=
\arg\max_{\theta\in\mathcal{V}_{K,\mathrm{dev}}}
\widehat{J}_{D_{\mathrm{dev}}}(\theta).
\]
The final reported test performance is then
\(\widehat{J}_{D_{\mathrm{test}}}(\hat{\theta}_{\mathrm{AS}})\).

\section{Proposed Method: \proposed}
\label{sec:method}

We now present \proposed, an iterative framework for automatic harness optimization of LLM agents. We begin with an overview of the workflow and then describe each key component in detail.

\paragraph{Overview.}
We formulate harness optimization as an offline learning problem. This formulation reflects the practical setting in which agent harnesses are typically tuned during development before being deployed to production. Since training may involve many tasks and rollout-based evaluation is expensive, we adopt a mini-batch training paradigm, following standard machine learning practice and prior work on automatic prompt optimization~\cite{DBLP:conf/emnlp/PryzantI0L0023, agrawal2026gepa}.

\begin{figure}[!t]  
\centering
    \includegraphics[width=1\linewidth]{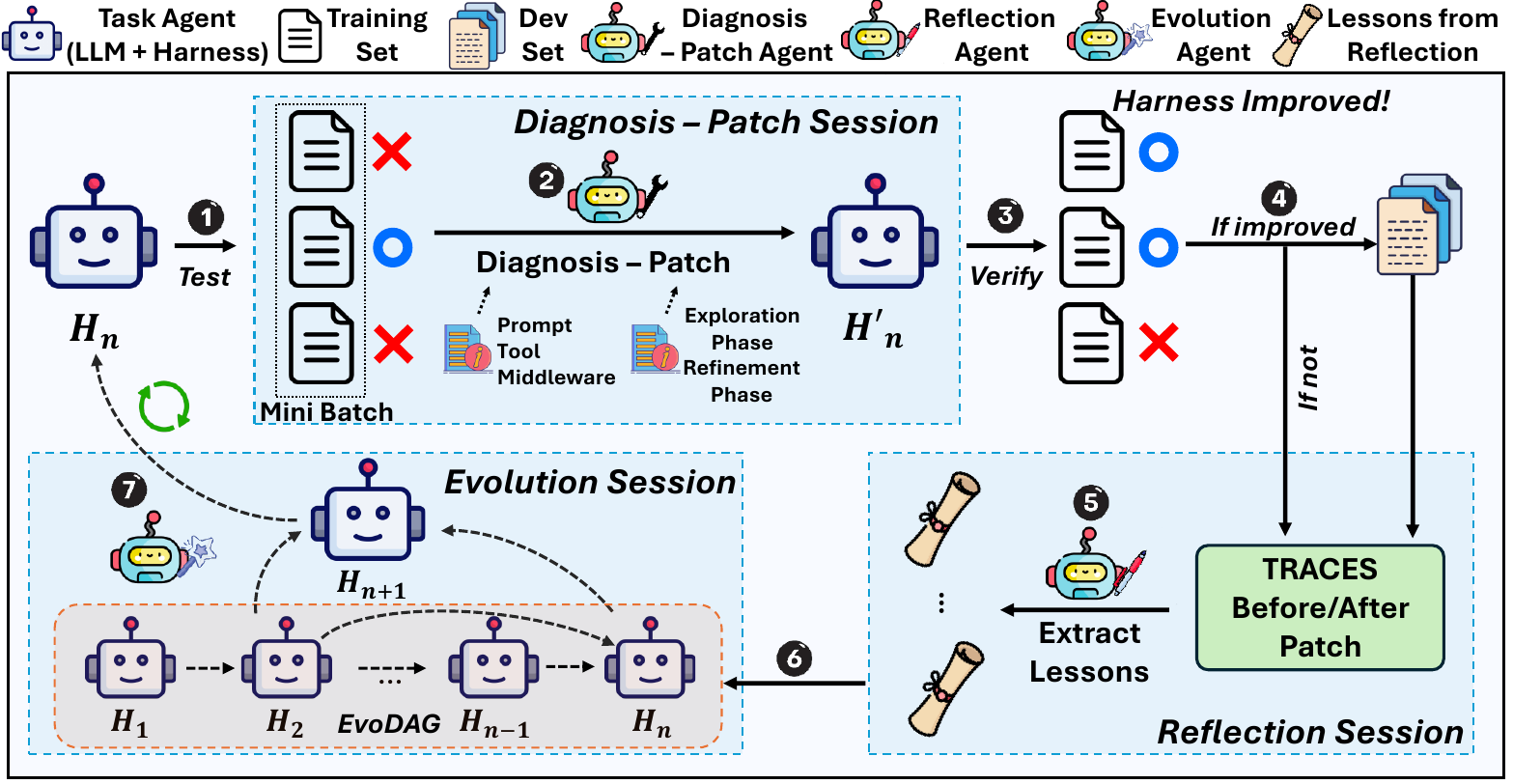} 
    \caption{\textbf{Overview of \proposed}. The iterative optimization loop: the current harness is tested on a mini-batch, diagnosed and patched across harness components, verified for improvement, and then reflected upon to extract lessons into the EvoDAG, which guides the evolution of the next harness.}
    \label{fig:main_figure}
\end{figure}

Specifically, we partition the task set $\mathcal{X}$ into training, development, and test splits, denoted by \(D_{\mathrm{train}}\), \(D_{\mathrm{dev}}\), and \(D_{\mathrm{test}}\), respectively. As illustrated in \autoref{fig:main_figure}, each iteration \(n\) begins by evaluating the current harness \(H_n\), parameterized by \(\theta_n\), on a mini-batch \(B_n \subset D_{\mathrm{train}}\) (\bcirc{1}). The workflow then enters the \textbf{Diagnosis--Patch Session}, which analyzes failed execution traces in the mini-batch and derives a structured patch \(\Delta\theta_n\), producing an updated harness \(H_n^\prime = H_n + \Delta\theta_n\) (\bcirc{2}). The patched harness is verified on the same mini-batch (\bcirc{3}); we regard the patch as a mini-batch improvement if $\widehat{J}_{B_n}(H^\prime_n) > \widehat{J}_{B_n}(H_n)$. When this criterion is satisfied, we further evaluate \(H_n^\prime\) on \(D_{\mathrm{dev}}\) to estimate whether the update generalizes beyond the observed mini-batch (\bcirc{4}).

Regardless of whether the patch is accepted, the \textbf{Reflection Session} compares pre- and post-patch traces; records fixed, regressed, still-failing, and still-passing cases (\bcirc{5}); and stores the resulting lessons and scores in \textbf{EvoDAG}, a directed acyclic graph that tracks the evolution history of harness updates (\bcirc{6}). The \textbf{Evolution Session} then uses EvoDAG to propose the next candidate harness \(H_{n+1}\) (\bcirc{7}). After the rollout budget \(K\) is exhausted, \proposed{} returns the candidate with the highest empirical development-set score among all candidates evaluated on \(D_{\mathrm{dev}}\). The selected harness is then evaluated once on the held-out test set and is not further modified using test feedback.

This workflow mirrors standard mini-batch learning, but adapts it to textual harness optimization. The Diagnosis--Patch--Verification steps (\bcirc{2} and \bcirc{3}) play the role of backpropagation under textual gradients: because textual error signals are not automatically checked like numerical gradients, each update requires explicit hypothesis generation, harness intervention, and empirical verification. EvoDAG and the Evolution Session play an optimizer-like role by accumulating reflection signals to perform history-aware harness evolution. We detail this analogy and its discrepancies in \autoref{app:autosaddler-ml-mapping}.

\paragraph{Diagnosis-Patch Session.} 
During this session, the execution traces from the mini-batch, including both successful and failed runs, are passed to the \textit{Diagnosis-Patch Agent} for failure diagnosis and patch generation. We do \emph{not} separate diagnosis from patch generation, allowing the agent to fully leverage the contextual information gathered during diagnosis when producing patches. Our diagnosis procedure is designed to be principled and evidence-based. In addition to each failure trace, we provide the agent with the harness codebase $\theta_n$ and structured guidance that enables it to progressively retrieve relevant trace details, thereby mitigating the long-context challenges posed by long agent trajectories. Based on this evidence, the agent identifies the suspected root cause, considers alternative hypotheses, and proposes recommendations for patch generation.

Patch generation is likewise structured. Specifically, the agent produces a \textit{structured patch} $\Delta \theta_n$ that targets one or more harness components. Rather than allowing unconstrained edits over the full harness codebase, we expose only the source files that implement the harness's functional logic, while prohibiting access to code related to evaluation or benchmark data. We further organize the patch space into three categories: \textit{Prompt}, \textit{Tool}, and \textit{Middleware}, each corresponding to a distinct layer of the agent harness. \autoref{tab:patch-taxonomy} enumerates the subtypes of each category and provides their descriptions.

We further divide these patch types into two higher-level groups: \textbf{Capability Patches} and \textbf{Steering Patches}. Capability Patches modify executable code or orchestration logic, whereas Steering Patches consist of textual edits that leave the underlying code unchanged, including modifications to prompts, tool descriptions, and hook reminder texts. Since these two patch groups may affect the harness in different ways, we introduce \textbf{Phased Patch Scheduling}, analogous to learning-rate scheduling in standard model training. In this schedule, optimization begins with a Capability Patch phase and then transitions to a Steering Patch phase. Additional details on Phased Patch Scheduling are provided in \autoref{app:autosaddler-ml-mapping}, while the diagnosis and patch-generation instructions are described in \autoref{sec:instructions_used_in_autosaddler}.

\begin{table}[t]
\fontsize{8}{9}\selectfont
\centering\caption{Adopted patch categories, with subtypes labeled as Capability (C) or Steering (S) Patch.}
\label{tab:patch-taxonomy}
\begin{tabular}{@{}clp{8cm}@{}}
\toprule
\textbf{Category} & \textbf{Subtype} & \textbf{Description} \\
\midrule
\multirow{2}{*}{\makecell[c]{Prompt\\Patch}}
  & Prompt Rule Addition (S) & Add or modify behavioral rules in the system prompt. \\
  & Prompt Rule Modification (S) & Revise existing prompt rules to resolve conflicts or sharpen guidance. \\
\midrule
\multirow{4}{*}{\makecell[c]{Tool\\Patch}}
  & New Tool Addition (C) & Add a new tool when no existing tool supports the required action. \\
  & Argument Modification (C) & Add or fix tool parameters to enhance filtering or selection capabilities. \\
  & Implementation Fix (C) & Fix bugs or extend internal tool functionality to produce correct results. \\
  & Tool Description Fix (S) & Modify tool docstrings to avoid tool misuse. \\
\midrule
\multirow{3}{*}{\makecell[c]{Middleware\\Patch}}
  & PreToolUse Hook (S) & Inject a just-in-time reminder before a specific tool call. \\
  & Infrastructure Change (C) & Modify agent configuration, iteration budget, or environment settings. \\
  & Agent Loop Logic Change (C) & Add preprocessing steps or budget reminders to the agent loop. \\
\bottomrule
\end{tabular}
\end{table}

\paragraph{Reflection Session.}
After evaluating the patched harness $H'_n$ on the same mini-batch, and optionally on the development set, we enter the \textit{Reflection Session} to extract lessons from the current harness update. Specifically, the \textit{Reflection Agent} is instructed to compare agent performance before and after applying the patch on the mini-batch. The resulting outcomes are grouped into four categories: \textit{``fixed''}, \textit{``regressed''}, \textit{``still-failing''}, and \textit{``still-passing''}. For each category, we provide targeted self-reflection questions to elicit insights into why the patch was effective, which failure patterns it addressed, why regressions occurred, why the patch was insufficient, or whether the patch had any effect at all. The agent may also inspect trace details during reflection. Furthermore, if the patch triggers an extended evaluation on the development set, we additionally prompt reflection on the development-set performance, with particular emphasis on whether and how the patch generalizes beyond the mini-batch. These lessons, together with the patch description, mini-batch results, and evaluation metric results $\mu(H'_n)$, are stored as node-level attributes in the EvoDAG to facilitate harness evolution. More details on the reflection instructions can be found in \autoref{sec:instructions_used_in_autosaddler}.

\paragraph{Evolution Session.}
The EvoDAG is a directed acyclic graph $\mathcal{G} = (V, E)$ that serves as the cumulative memory of the optimization process. Each node $v_n \in V$ corresponds to a previously explored harness and is annotated with its associated lessons and performance signals. Each directed edge $e \in E$ represents the diff $\Delta \theta$ between a parent harness and its descendant.

During the Evolution Session, the \textit{Evolution Agent} consults the full EvoDAG to synthesize a new harness $H_{n+1}$. Rather than continuing solely from $H_n^\prime$, the agent can compose elements from any subset of previously explored harnesses, guided by the accumulated lessons stored in the graph. This merge operation is analogous to evolutionary search, enabling the framework to escape local optima by recombining successful components across different lineages in the optimization history. The resulting harness $H_{n+1}$ is then evaluated on a fresh mini-batch in the next iteration. Further details of the evolution instructions are provided in \autoref{sec:instructions_used_in_autosaddler}.

\paragraph{Design Rationale.}
Overall, \proposed{} is designed around three requirements for
effective automatic harness optimization: \emph{in-depth diagnosis},
\emph{structured intervention}, and
\emph{generalization-aware selection}. The Diagnosis-Patch Session supports in-depth diagnosis by grounding
updates in explicit analysis of long-horizon execution traces and the
harness codebase, enabling deep debugging rather than shallow
reflection. The patch taxonomy and phased schedule support structured
intervention by restricting optimization to targeted changes over
prompts, tools, and middleware, rather than unconstrained editing.
The Reflection and Evolution Sessions support generalization-aware
selection by combining mini-batch verification, optional
development-set evaluation, and EvoDAG-based memory, so that retained
updates are more likely to generalize beyond a single trajectory. Together, these components turn harness optimization into an iterative process of diagnosing failures, applying targeted updates, and retaining changes that are more likely to generalize.

\section{Experiments}
\label{sec:experiments}

\subsection{Experimental Setup}

\paragraph{Benchmarks and Base Harness Systems.}
We evaluate \proposed{} on three diverse benchmarks: GAIA2~\cite{GAIA2}, SWE-Bench Pro (SBP)~\cite{SWE-Bench-Pro}, and Terminal-Bench 2.0 (TB2)~\cite{TerminalBench2}. GAIA2 evaluates general-purpose assistant capabilities in a simulated smartphone environment spanning 10 distinct \textit{Universes}, each emulating a persona's daily digital life. We use the default ReAct-based agent provided by GAIA2 as the base harness. SBP targets enterprise-scale software engineering tasks drawn from popular repositories, for which we use SWE-agent~\cite{sweagent} as the base harness. Finally, TB2 comprises 89 realistic tasks across domains including system administration, machine learning, and cybersecurity; for TB2, we adopt Terminus 2 as the base harness.

\paragraph{Implementation and Baseline Methods.}
We implement the three agents in \proposed{} (i.e., the \textit{Diagnosis-Patch Agent}, \textit{Reflection Agent}, and \textit{Evolution Agent}) using Claude Agent SDK (CA-SDK)~\cite{CCSDK}. Detailed agent instructions are provided in \autoref{sec:instructions_used_in_autosaddler}, and EvoDAG implementation details are given in \autoref{app:more_details_exp_setup}. We compare \proposed{} against GEPA~\cite{agrawal2026gepa} and Meta-Harness~\cite{lee2026meta}. GEPA is a prompt-centric baseline that iteratively optimizes system prompts. By contrast, Meta-Harness shifts the optimization target from prompts to the agent harness, using a minimal system design and adopting CA-SDK for end-to-end system optimization. Since only TB2 was evaluated by Meta-Harness, we adapt GEPA to GAIA2 and TB2, and Meta-Harness to GAIA2; adaptation details are provided in \autoref{app:more_details_exp_setup}.

\paragraph{Experimental Protocol and Data Splits.}
Unless otherwise noted, all optimization methods use \texttt{Claude Opus 4.6} as the underlying LLM with default endpoint settings. To evaluate generalization, we ensure that the training, development, and test sets contain tasks from distinct task groups. For example, in SBP, the training set consists of tasks from \texttt{qutebrowser}; the development set contains tasks from \texttt{Vuls} and \texttt{NodeBB}; and the test set comprises tasks from \texttt{Ansible}, \texttt{Flipt}, and \texttt{Element-web}. This split enables us to evaluate generalization across repositories. Additional details on the data splits for all three benchmarks are provided in \autoref{app:more_details_exp_setup}. Unless otherwise noted, for each optimization method, we perform a single evolution run on the training and development sets, and conduct \textit{three} repeated runs when evaluating on the test sets.\footnote{Because Meta-Harness does not natively support a train--dev split, we provide it with the union of the training and development sets to ensure comparable data exposure.} We report \textit{Pass@1} success rates as the evaluation metric.

\subsection{Main Results}
\label{sec:main_results}

\begin{table}[tbp]
\centering
\small
\setlength{\tabcolsep}{4pt} 
\caption{Test-set Pass@1 results on GAIA2, reported as mean $\pm$ standard deviation over three runs. Values in parentheses indicate the number of tasks; boldface denotes the best result.}
\label{tab:eval_results_gaia2}
\sisetup{
  separate-uncertainty = true,
  table-format         = 2.1(2),
  detect-weight        = true,
  detect-inline-weight = math,
}
\begin{tabular}{lSSSS}
\toprule
\multicolumn{1}{c}{\multirow{2}{*}{\textbf{Harness (Type)}}}
 & \multicolumn{3}{c}{\textbf{GAIA2 Universe (Pass@1)}}
 & \multicolumn{1}{c}{\multirow{2}{*}{\textbf{Avg.}}} \\
\cmidrule(lr){2-4}
 & {\textbf{21 (107)}} & {\textbf{22 (112)}} & {\textbf{27 (81)}} & \\
\midrule
Default Agent (Manual)                      & 54.8(48)           & 51.5(49)           & 52.7(43)           & 53.0(15)           \\
GEPA (Auto)                                 & 60.1(39)           & 47.9(34)           & 56.4(7)            & 54.6(25)           \\
Meta-Harness (Auto)                         & 53.0(11)           & 51.5(52)           & 56.0(7)            & 53.2(22)           \\
\textbf{\proposed{} (Auto)}                 & \bfseries 61.4(24) & \bfseries 60.7(24) & \bfseries 64.6(31) & \bfseries 62.0(12) \\
\midrule
w/o In-depth Diagnosis (Auto)               & 56.7(42)           & 57.1(39)           & 60.1(50)           & 57.8(38)           \\
w/o Structured Intervention (Auto)          & 58.9(25)           & 53.3(76)           & 59.3(33)           & 56.9(38)           \\
w/o Generalization-Aware Selection (Auto)   & 53.3(19)           & 44.9(67)           & 54.7(40)           & 50.6(40)           \\
\bottomrule
\end{tabular}
\end{table}

\begin{table}[tbp]
\centering
\footnotesize
\setlength{\tabcolsep}{3pt}
\caption{Test-set Pass@1 results on SWE-Bench Pro and Terminal-Bench 2.0, reported as mean $\pm$ standard deviation over three runs. Parentheses indicate the number of tasks; bold denotes the best result per column.}
\label{tab:eval_results_combined}
\sisetup{
  separate-uncertainty    = true,
  retain-zero-uncertainty = true,
  table-format            = 2.1(2),
  detect-weight           = true,
  detect-inline-weight    = math,
}

\begin{tabular}{@{}p{0.60\textwidth}@{\hspace{2em}}p{0.32\textwidth}@{}}

\begin{minipage}[t]{\linewidth}
\centering
\vspace{0.2em}
\textbf{SWE-Bench Pro (Pass@1)}
\vspace{0.25em}

\resizebox{\linewidth}{!}{%
\begin{tabular}{lSSSS}
\toprule
\textbf{Harness (Type)}
 & {\textbf{\texttt{Ansible} (96)}} 
 & {\textbf{\texttt{Flipt} (85)}}
 & {\textbf{\texttt{Element-web} (56)}}
 & \textbf{Avg.} \\
\midrule
SWE-agent (Manual)          & 40.6(19)           & 31.0(35)           & 41.1(17)           & 37.3(48) \\
GEPA (Auto)                 & 50.0(13)           & 32.2(14)           & \bfseries 45.2(06)           & 42.5(12) \\
Meta-Harness (Auto)         & 36.9(29)           & 31.3(17)           & 38.7(08)           & 35.3(20) \\
\textbf{\proposed{} (Auto)} & \bfseries 58.0(18) & \bfseries 36.5(18) &  43.5(16) & \bfseries 46.9(18) \\
\bottomrule
\end{tabular}%
}

\end{minipage}
&
\begin{minipage}[t]{\linewidth}
\centering
\textbf{Terminal-Bench 2.0  (Pass@1)}
\vspace{0.25em}

\resizebox{\linewidth}{!}{%
\begin{tabular}{lS}
\toprule
\textbf{Harness (Type)}
 & {\textbf{Test-Split (40)}} \\
\midrule
Terminus 2 (Manual)         & 40.0(0)  \\
Terminus KIRA (Manual)      & 47.5(25) \\
GEPA (Auto)                 & 42.5(25) \\
Meta-Harness (Auto)         & 43.3(58) \\
\proposed{} (Iter2) (Auto)  & 45.0(0)  \\
\textbf{\proposed{} (Iter34) (Auto)} & \bfseries 50.0(0) \\
\bottomrule
\end{tabular}%
}
\end{minipage}

\end{tabular}

\end{table}

We report Pass@1 success rates on the test sets for GAIA2 in \autoref{tab:eval_results_gaia2}, and for SBP and TB2 in \autoref{tab:eval_results_combined}. Two main observations are made. First, \textbf{\proposed{} discovers more effective harnesses than the corresponding base harnesses across all three benchmarks.} Specifically, \proposed{} improves over the default agent on GAIA2 by \textbf{+9.0\,pp} ($53.0\% \rightarrow 62.0\%$), over SWE-agent on SBP by \textbf{+8.4\,pp} ($37.3\% \rightarrow 46.9\%$), and over Terminus 2 on TB2 by \textbf{+10.0\,pp} ($40.0\% \rightarrow 50.0\%$). Second, \textbf{\proposed{} outperforms the automated baselines GEPA and Meta-Harness.} Compared with the strongest automated baseline on each benchmark, \proposed{} achieves gains of \textbf{+7.4\,pp} on GAIA2 ($54.6\%$ vs.\ $62.0\%$), \textbf{+6.2\,pp} on SBP ($42.5\%$ vs.\ $46.9\%$), and \textbf{+4.4\,pp} on TB2 ($43.3\%$ vs.\ $50.0\%$). On TB2, \proposed{} also surpasses the manually expert-tuned harness, \textit{Terminus KIRA}~\cite{terminuskira2026}, by \textbf{+2.5\,pp} ($47.5\%$ vs.\ $50.0\%$).

We further verify robustness to optimization stochasticity and training-distribution shift. 
An independent \proposed{} run on GAIA2 achieves $58.6\%$ Pass@1, remaining well above the base harness and rerun baselines (\autoref{app:optimization_stochasticity}), while optimization on a different training universe achieves $57.4\%$, a \textbf{+5.9\,pp} gain over the base harness (\autoref{app:training_distribution_robustness}).


Beyond cross-task-group evaluation, we further analyze \textbf{cross-model transferability}. Specifically, we replace the underlying LLM of the GAIA2 task agent, switching from \texttt{Opus 4.6} to \texttt{Haiku 4.5}, while retaining the harnesses optimized with \texttt{Opus 4.6}. As shown in \autoref{app:appendix_cross_model}, \proposed{} still improves over the base harness by \textbf{+5.6\,pp}, suggesting potential transferability across models.

To analyze optimization efficiency, we compare the optimization trajectories of \proposed{} and the automated baselines on GAIA2 in \autoref{fig:optimization_comparison}; a visualization of the full search trajectory of \proposed{} is provided in \autoref{app:search_trajectory_visualization}. We find that \textbf{\proposed{} achieves higher performance with substantially fewer task-agent rollouts.} Specifically, \proposed{} reaches $72.3\%$ development accuracy with only ${\sim}1{,}000$ rollouts, whereas GEPA and Meta-Harness saturate at 64.6\% and 61.5\%, respectively, despite consuming ${\sim}$2{,}800 task executions (\autoref{fig:compute_efficiency}). The contrast is even sharper when efficiency is measured by the number of rollouts leveraged for learning (\autoref{fig:learning_efficiency}): \proposed{} reaches its best dev-set score after consuming $147$ rollouts, $\mathbf{{\sim}10}\boldsymbol{\times}$ fewer than Meta-Harness ($1{,}400$ rollouts). Similar efficiency gains are observed on TB2, as shown in \autoref{appendix:optimization_efficiency_other_benchmarks}.
We further characterize end-to-end optimization cost in \autoref{app:full_optimization_cost}. Although \proposed{} incurs moderately higher optimizer-side monetary cost per patch, its selective evaluation strategy reaches $67.7\%$ development accuracy with only $391$ task-agent rollouts, already exceeding Meta-Harness's $61.5\%$ peak after $1{,}400$ rollouts.

\subsection{Ablation Studies and Analysis}

We next present ablation studies on GAIA2, focusing on the three key design principles underlying \proposed{}: \emph{in-depth diagnosis}, \emph{structured intervention}, and \emph{generalization-aware selection}.

\paragraph{RQ1: Does in-depth diagnosis enable superior root-cause identification?} 

We first conduct an ablation study to assess whether in-depth diagnosis improves root-cause identification and ultimately contributes to overall performance gains. Specifically, we evaluate a ``\textit{w/o in-depth diagnosis}'' variant that replaces CA-SDK-based diagnosis with a shallow diagnostic baseline: a single LLM call receives the execution trace and evaluation results, and infers the failure reason, a strategy commonly used in automatic prompt optimization pipelines for failure reflection~\cite{DBLP:conf/emnlp/PryzantI0L0023, agrawal2026gepa}. The inferred failure reason is then passed back to CA-SDK for patch generation. In contrast, \proposed{}'s in-depth diagnosis actively explores both execution traces and source code to investigate failures.

As shown in \autoref{tab:eval_results_gaia2}, removing in-depth diagnosis substantially degrades test-set performance on GAIA2, \textbf{reducing Pass@1 from 62.0 to 57.8}. The benefit of in-depth diagnosis is further supported by the detailed file-access and tool-call analysis in~\autoref{app:more_details_RQ1}: compared with the patch-only session, the combined diagnosis--patch session invokes, on average, 6.2 additional tool calls and 5.8 additional file accesses per optimization step. We also quantitatively compare the cumulative number of accepted patches during training under the two settings, where an accepted patch is defined as one that improves performance on the same mini-batch; see~\autoref{fig:rq1_accepted_patches} in~\autoref{app:more_details_RQ1}. In-depth diagnosis consistently yields more accepted patches. By the end of Epoch~1 (Iteration~25), the gap is substantial (\textbf{13} vs.\ \textbf{5}), and this advantage is maintained throughout Epoch~2 (Iterations~26--50). Finally, we provide several qualitative case studies in~\autoref{app:more_details_RQ1} that illustrate how shallow diagnosis and \proposed{}'s in-depth diagnosis differ in identifying root causes.

\paragraph{RQ2: How does structured intervention shape patch diversity and effectiveness?}

\begin{figure}[!h]
\centering
    \begin{subfigure}[b]{0.253\linewidth}
        \centering
        \includegraphics[width=\linewidth]{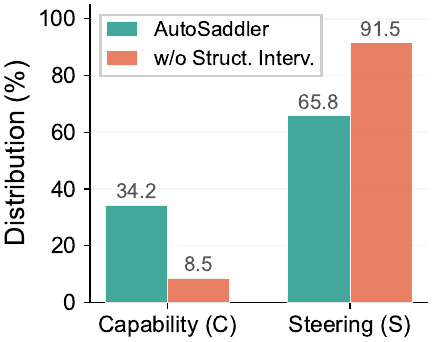}
        \caption{Capability vs.\ Steering.}
        \label{fig:patch_diversity_cs}
    \end{subfigure}
    \hfill
    \begin{subfigure}[b]{0.348\linewidth}
        \centering
        \includegraphics[width=\linewidth]{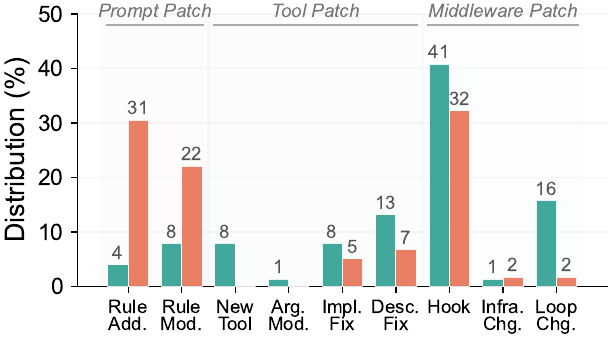}
        \caption{Patch subtype distribution.}
        \label{fig:patch_diversity_subtype}
    \end{subfigure}
    \hfill
    \begin{subfigure}[b]{0.348\linewidth}
        \centering
        \includegraphics[width=\linewidth]{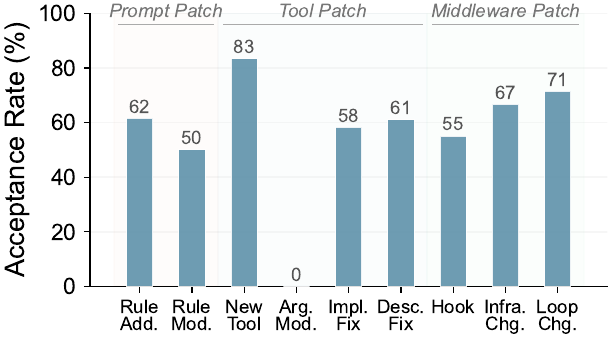}
        \caption{Acceptance rate by subtype.}
        \label{fig:patch_acceptance_subtype}
    \end{subfigure}
    \caption{
        \textbf{Patch type distribution and acceptance: \proposed{} vs.\ \emph{w/o Structured\ Intervention.}}
        Without structural intervention, patches collapse onto Steering (91.5\%), while \proposed{} produces a balanced mix spanning Prompt, Tool, and Middleware edits.
    }
    \label{fig:patch_diversity}
\end{figure}

To evaluate the effectiveness of structured intervention, we consider an ablated setting, \textit{w/o Structured Intervention}, in which we remove the proposed patch taxonomy and phased patch scheduling. This ablation leaves the agent to perform unconstrained edits without explicit search-space boundaries (as in Meta-Harness). As shown in \autoref{tab:eval_results_gaia2}, removing structured patching substantially reduces Pass@1 on the GAIA2 test set, \textbf{from 62.0\% to 56.9\%}. This result suggests that framing harness optimization as a targeted structural search is critical for achieving meaningful performance gains.
Fine-grained ablations in \autoref{app:fine_grained_ablations} further show that removing only Phased Patch Scheduling reduces Pass@1 from $60.7\%$ to $54.8\%$, while removing the full structured intervention further reduces it to $53.3\%$.

Further analysis shows that the performance degradation in the \textit{w/o Structured Intervention} setting stems from biased patch exploration and reduced patch diversity, despite allowing unconstrained patch edits. As shown in \autoref{fig:patch_diversity_cs} and \autoref{fig:patch_diversity_subtype}, without structured intervention, the system becomes heavily concentrated on Steering patches ($91.5\%$), which largely correspond to straightforward textual edits, while rarely exploring higher-value interventions such as infrastructure or tool improvements. The varying effectiveness of different patch types is further supported by the acceptance-rate breakdown across patch subtypes in \autoref{fig:patch_acceptance_subtype}. Capability-centric, non-prompt-layer patches, i.e., \textit{New Tool} ($83\%$), \textit{Loop Change} ($71\%$), and \textit{Infra Change} ($67\%$), achieve the highest acceptance rates. However, under \textit{w/o Structured Intervention}, these critical patch types account for only \textbf{4\%} of generated patches; by contrast, \proposed{} increases their share to over \textbf{25\%} through structured patching. Representative patches from these categories discovered by \proposed{} during evolution are provided in \autoref{app:more_details_RQ2}. Together, these results indicate that structured intervention encourages the agent to move beyond prompt-layer edits and explore capability-level improvements.
Moreover, \autoref{app:patch_durability} shows that Capability Patches achieve a comparable fix rate to Steering Patches ($55\%$ vs.\ $58\%$) while inducing substantially fewer regressions ($8\%$ vs.\ $17\%$), suggesting more durable updates.

\paragraph{RQ3: How does generalization-aware selection prevent overfitting?}

\begin{figure}[!h]
\centering
    \begin{subfigure}[b]{0.32\linewidth}
        \centering
        \includegraphics[width=\linewidth]{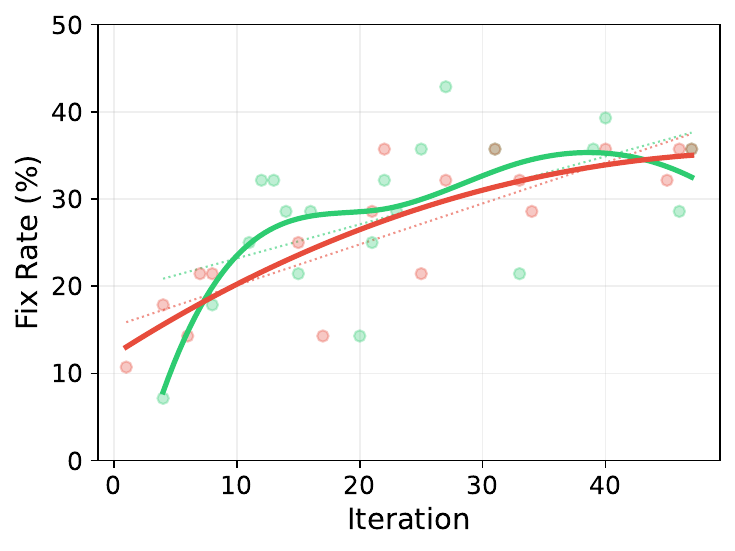}
        \caption{Fix Rate (Base-Failed).}
        \label{fig:rq3_dev_fix_rate}
    \end{subfigure}
    \hfill
    \begin{subfigure}[b]{0.32\linewidth}
        \centering
        \includegraphics[width=\linewidth]{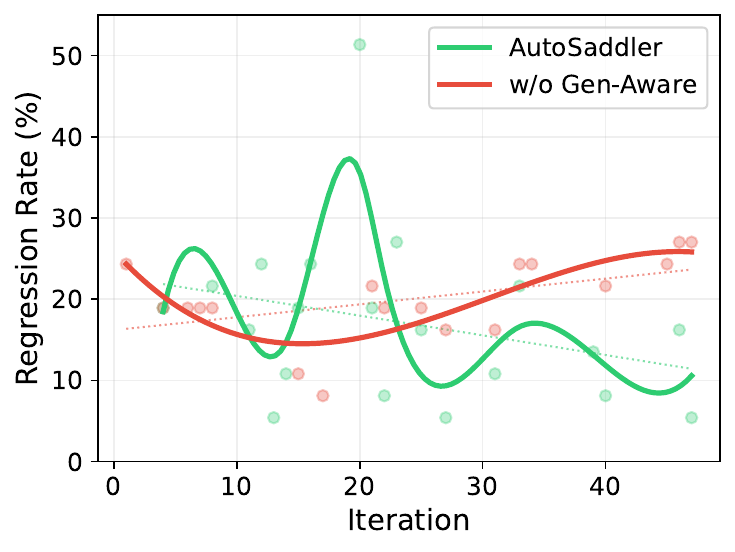}
        \caption{Regression Rate (Base-Passed).}
        \label{fig:rq3_dev_regression_rate}
    \end{subfigure}
    \hfill
    \begin{subfigure}[b]{0.32\linewidth}
        \centering
        \includegraphics[width=\linewidth]{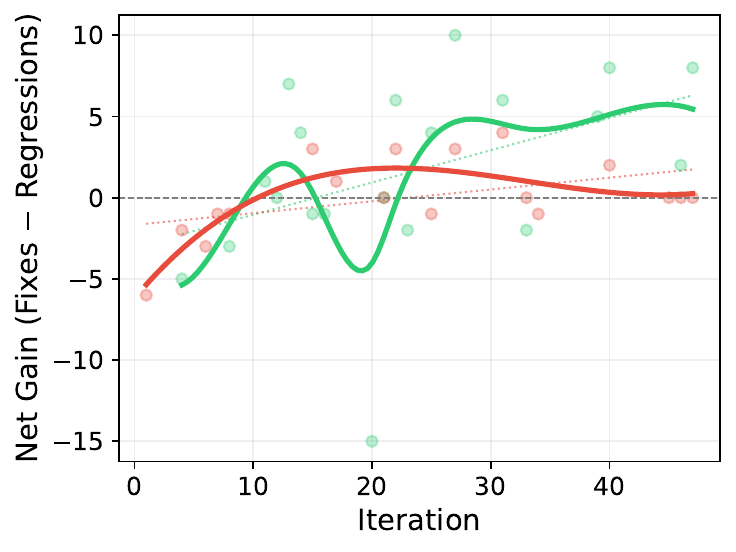}
        \caption{Dev-Set Net Gain.}
        \label{fig:rq3_dev_net_gain}
    \end{subfigure}
    \caption{
        \textbf{Performance comparison on the dev-set across iterations.}
        \proposed{} maintains a lower regression rate while achieving comparable fix rates to the \emph{w/o Generalization-Aware Selection} ablation, yielding a consistently positive net gain that the ablation fails to sustain.
    }
    \label{fig:rq3_gen_aware}
\end{figure}

Finally, we conduct an ablation study to examine how generalization-aware optimization mitigates overfitting to the training set and reduces regressions on unseen scenarios. Specifically, the "\textit{w/o Generalization-Aware Selection}" ablation removes both the reflection session and dev-set evaluation, forcing optimization to rely solely on training-set execution outcomes. As shown in \autoref{tab:eval_results_gaia2}, removing this component substantially degrades GAIA2 test-set performance, \textbf{reducing Pass@1 from 62.0\% to 50.6\%}. This is the largest performance drop among all component ablations.
Fine-grained ablations in \autoref{app:fine_grained_ablations} further show that removing dev-set filtering reduces Pass@1 from $60.7\%$ to $50.0\%$, while additionally removing Reflection with EvoDAG further reduces it to $44.9\%$.

To quantify how this mechanism affects dev-set behavior, \autoref{fig:rq3_gen_aware} tracks both settings on the dev-set and decomposes net performance into the \textit{fix rate}, defined as the success rate on scenarios failed by the initial base harness, and the \textit{regression rate}, defined as the failure rate on scenarios passed by the base harness. Notably, both settings achieve similar fix rates (panel~a), indicating that the performance gap is not primarily driven by differences in problem-solving capability. Instead, the key divergence lies in the regression rate (panel~b). Although both curves fluctuate during optimization, \proposed{} exhibits an overall decreasing regression trend (\textbf{-0.24 pp/iter}), whereas the ablation shows an increasing trend (\textbf{+0.16 pp/iter}).

We further examine the regression spike observed for the ablation at Iteration~21 (panel~b) ($8\% \rightarrow 22\%$). At Iteration~20, the ablation introduces a new tool, \texttt{send\_progress\_message\_to\_user}, and modifies the hook for the widely used \texttt{send\_message\_to\_user} tool to forcibly redirect the agent to this new tool. Without reflection to assess collateral damage, this overly broad patch is retained, disrupting agent behavior across many unrelated development scenarios. Interestingly, the same failure pattern---an overly broad hook on a high-frequency tool that acts beyond its intended scope---also appears in \proposed{} at Iteration~4 (see \autoref{fig:rq3_reflection_lessons}b and the extended discussion in \autoref{app:more_details_RQ3}). However, because of the reflection mechanism in \proposed{}, such regression-inducing patches are blocked. This direct contrast demonstrates that generalization-aware selection helps sustain performance gains on unseen scenarios by identifying and filtering over-scoped patches.

Lastly, \autoref{fig:rq3_dev_net_gain} shows the resulting net gain, indicating that \proposed{} consistently outpaces the ablation after Iteration~20. This suggests that \proposed{} can recover from early-iteration regression challenges and achieve steadier improvements in later iterations by abstracting generalizable principles from specific optimization lessons. Additional discussion of principle generalization and qualitative case studies is provided in \autoref{app:more_details_RQ3}.

\section{Conclusion}

In this work, we introduced \proposed, an automatic harness optimization framework that formulates harness improvement for LLM agents as an offline learning problem over execution traces. \proposed{} combines in-depth failure diagnosis, structured patch generation across prompts, tools, and middleware, and generalization-aware update selection via validation and EvoDAG-based evolution. As a result, it produces durable harness updates rather than trajectory-specific fixes. On GAIA2, SWE-Bench Pro, and Terminal-Bench 2.0, \proposed{} consistently improves over the corresponding base harnesses by \textbf{9.0, 9.6, and 10.0} percentage points, respectively, and outperforms the strongest automated baseline on each benchmark by \textbf{7.4, 4.4, and 6.7} points. These results show that effective harness optimization requires deep debugging, targeted interventions, and explicit selection for generalization, positioning automatic harness optimization as a promising direction for building more performant and reliable long-horizon agent systems.

\section{Acknowledgments}

We thank the anonymous reviewers for their constructive comments and suggestions, which helped improve the clarity and quality of this work. We are grateful to Bo Qiao for his generous help with setting up the computing servers used for our experiments. We further thank the creators and maintainers of GAIA2, SWE-Bench Pro, Terminal-Bench 2.0, SWE-agent, Terminus 2, GEPA, and Meta-Harness for making their benchmarks, systems, and tools available, which enabled the experiments in this work.  
\bibliographystyle{plain}
\bibliography{bibliography}

\newpage
\appendix

\section{AutoSaddler as Mini-Batch Learning over Textual Harness Parameters}
\label{app:autosaddler-ml-mapping}

This appendix clarifies how the \proposed{} loop in \autoref{fig:main_figure} relates to the standard mini-batch training loop. The analogy is useful because \proposed{} preserves the outer structure of mini-batch learning: it repeatedly evaluates a candidate system on a mini-batch, extracts an update signal, checks whether the update generalizes, and uses accumulated optimization history to propose the next candidate. However, the analogy should not be taken literally. Unlike neural parameters, harness parameters are textual, executable, and non-differentiable. \proposed{} therefore replaces differentiable gradient computation with a trace-grounded symbolic optimization procedure.

\paragraph{Mapping to traditional mini-batch training.}
\autoref{tab:ml-autosaddler-mapping} summarizes the correspondence. We denote the stages of traditional mini-batch training as Step~1--Step~7, and refer to \proposed{} stages using the circled labels from \autoref{fig:main_figure}.

\begin{table}[t]
\centering
\footnotesize
\begin{tabular}{p{0.3\linewidth} p{0.3\linewidth} p{0.3\linewidth}}
\toprule
\textbf{Traditional Mini-batch Training} & \textbf{\proposed{} Stage} & \textbf{Interpretation} \\
\midrule
Step~1: Sample mini-batch & \bcirc{1} Sample mini-batch for evaluating current harness & Same role \\
\addlinespace
Step~2: Run forward pass & \bcirc{1} Execute the current harness & The agent rollout is the forward computation. It produces both a final answer and a trajectory. \\
\addlinespace
Step~3: Compute loss & \bcirc{1} Record outcomes and traces & The task outcome provides the sparse metric signal; the trace provides evidence needed to explain the outcome. \\
\addlinespace
Step~4: Backpropagate gradients & \bcirc{2} Diagnose and patch + \bcirc{3} verify on the same mini-batch & AutoSaddler has no numerical gradient; it constructs a textual update through hypothesis generation on root cause, intervention, and verification. \\
\addlinespace
Step~6: Validate/checkpoint & \bcirc{4} Evaluate on the dev-set & Dev-set evaluation checks whether the candidate improvement generalizes beyond the mini-batch. \\
\addlinespace
Step~5: Apply optimizer update & \bcirc{5} Reflect + \bcirc{6} store in EvoDAG + \bcirc{7} evolve the next harness & The committed update is not merely the raw patch. AutoSaddler updates optimizer state and uses it to synthesize the next candidate harness. \\
\addlinespace
Step~7: Repeat & Next iteration & The evolved harness is evaluated on a fresh mini-batch. \\
\bottomrule
\end{tabular}
\caption{Mapping between conventional mini-batch training and the AutoSaddler loop in \autoref{fig:main_figure}. AutoSaddler follows the outer structure of mini-batch learning but replaces differentiable gradient computation with trace-grounded symbolic optimization.}
\label{tab:ml-autosaddler-mapping}
\end{table}

\paragraph{Why backpropagation becomes diagnosis, patching, and verification.}
The central mismatch between the two settings lies in the analogue of backpropagation. In neural training, once the loss is computed, backpropagation provides a mathematically defined gradient. The update direction is therefore relatively well specified.

In \proposed{}, a failed rollout does not directly specify how the harness should change. The root cause may lie in the prompt, a tool interface, a tool implementation, a middleware hook, or the agent loop. The system must therefore infer a root-cause hypothesis from the trace, implement a targeted intervention, and empirically test whether the intervention improves behavior on the same mini-batch:
\begin{equation}
    \text{diagnosis}
    \;\longrightarrow\;
    \text{patch-as-intervention}
    \;\longrightarrow\;
    \text{same-batch verification}.
\end{equation}
This sequence plays the role of constructing a credible textual gradient. Importantly, the patch at this stage is not yet best understood as the final committed parameter update. Rather, it is an intervention used to test whether the inferred root cause is plausible. This distinction is important because numerical gradients are derived from a fixed mathematical operator, whereas textual gradients are inferred, semantic, and fallible. Same-batch verification is therefore necessary to avoid treating an untested explanation as a reliable update direction.

\paragraph{Why the optimizer update is reflection plus EvoDAG evolution.}
A second mismatch concerns the notion of an optimizer update. In ordinary training, the optimizer directly modifies numeric parameters. In AutoSaddler, the raw patch is only one candidate change. Before allowing this change to influence future search, the system must determine whether it is useful beyond the current mini-batch and what reusable lesson, if any, should be retained.

This is why the committed update is better understood as
\begin{equation}
    \text{dev-set validation}
    \;\longrightarrow\;
    \text{reflection}
    \;\longrightarrow\;
    \text{EvoDAG update}
    \;\longrightarrow\;
    \text{evolution of the next harness}.
\end{equation}
Reflection distills before/after evidence into reusable lessons, and EvoDAG stores these lessons together with patch descriptions and performance signals. The Evolution Session then uses this symbolic optimizer state to select, revise, or recombine components from previously explored harnesses when constructing the next candidate. In this sense, EvoDAG serves as a form of optimizer memory. Analogous to how momentum accumulates past gradient information to smooth noisy updates and steer future steps toward historically productive directions, EvoDAG accumulates past patches, outcomes, and reflections to guide harness evolution toward changes with evidence of generalizable benefit. Thus, AutoSaddler's update is history-aware and compositional, rather than a single local arithmetic step.

\paragraph{A note on ordering.}
The mapping above intentionally swaps the usual order of Step~5 and Step~6. In standard mini-batch training, one typically applies the optimizer update before validation or checkpointing. In \proposed{}, development-set validation precedes the final symbolic optimizer update because EvoDAG is part of the optimizer state. Once a misleading lesson is stored, it can influence future evolution. Validation therefore serves as a generalization gate before the system commits lessons to EvoDAG and evolves \(H_{n+1}\).

\paragraph{Algorithmic view.}
Algorithm~\ref{alg:autosaddler-minibatch-view} presents the same view algorithmically. The key point is that the patch produced in \bcirc{2}--\bcirc{3} functions as an intervention for testing an inferred textual gradient. The final update to the search trajectory occurs only after validation, reflection, and EvoDAG-based evolution.

\begin{algorithm}[t]
\small
\caption{AutoSaddler as mini-batch learning over harness parameters}
\label{alg:autosaddler-minibatch-view}
\begin{algorithmic}[1]
\Require Initial harness \(H_0\), training set \(D_{\mathrm{train}}\), development set \(D_{\mathrm{dev}}\), rollout budget \(K\)
\State Initialize EvoDAG \(G_0\) with \(H_0\)
\For{iteration \(n=0,1,2,\ldots\) until budget \(K\) is exhausted}
    \State Sample mini-batch \(B_n \subset D_{\mathrm{train}}\) \Comment{\autoref{fig:main_figure}: \bcirc{1}; Step~1}
    \State Execute \(H_n\) on \(B_n\), collecting outcomes and traces \(\tau_n\) \Comment{\bcirc{1}; Steps~2--3}
    \State Diagnose failures in \(\tau_n\) and infer root-cause hypotheses \Comment{\bcirc{2}; textual credit assignment}
    \State Generate targeted patch \(\Delta\theta_n\), yielding candidate \(H'_n = H_n + \Delta\theta_n\) \Comment{\bcirc{2}; patch as intervention}
    \State Re-evaluate \(H'_n\) on \(B_n\) \Comment{\bcirc{3}; verify textual gradient}
    \If{same-batch performance improves}
        \State Evaluate \(H'_n\) on \(D_{\mathrm{dev}}\) \Comment{\bcirc{4}; Step~6 generalization gate}
    \EndIf
    \State Reflect on before/after traces and evaluation results \Comment{\bcirc{5}; convert evidence into lessons}
    \State Update EvoDAG \(G_n\) with patch, metrics, and lessons to obtain \(G_{n+1}\) \Comment{\bcirc{6}; symbolic optimizer state}
    \State Evolve next harness \(H_{n+1}\) from \(G_{n+1}\) \Comment{\bcirc{7}; Step~5 committed update}
\EndFor
\State \Return best harness selected by development performance
\end{algorithmic}
\end{algorithm}

\paragraph{Phased patch scheduling.}
\label{subsec:patch-scheduling}
We divide the patch types in \autoref{tab:patch-taxonomy} into two categories based on the type of harness change they make. \textbf{Capability patches} modify executable code or orchestration logic, including tool implementations, tool arguments, infrastructure settings, and agent-loop logic. These patches can change what actions the agent is able to perform or how the harness executes those actions. \textbf{Steering patches} are textual edits that leave the underlying executable code unchanged, including modifications to prompts, tool descriptions, and hook reminder texts. These patches primarily refine how the agent selects among existing capabilities and follows task-specific constraints.

This distinction is analogous to, but not identical with, large versus small learning-rate steps in gradient-based optimization. Capability patches often behave like larger steps because they can introduce new functionality, change control flow, or alter the available action space. Steering patches often behave like smaller steps because they adjust the agent's behavior within an already established capability set. However, this analogy is only approximate: prompt or hook edits can sometimes produce large behavioral changes, especially when they affect high-frequency decisions. We therefore use the capability--steering distinction as a scheduling heuristic rather than a strict guarantee about effect size, and rely on verification and development-set validation to detect overly broad or regression-inducing changes.

Motivated by this analogy, \proposed{} adopts a two-phase schedule with transition point \(k\):
\begin{enumerate}[leftmargin=*,itemsep=2pt]
    \item \textbf{Exploration phase} (\(n \le k\)). The search prioritizes capability patches to address fundamental gaps in tooling, infrastructure, and agent-loop behavior.
    \item \textbf{Refinement phase} (\(n > k\)). The search switches to steering patches to refine behavior after the capability set has stabilized.
\end{enumerate}
The transition point \(k\) can be specified directly as an iteration count or implicitly through the number of training epochs:
\[
    k = E \cdot \left\lceil \frac{|\mathcal{D}_{\mathrm{train}}|}{B} \right\rceil,
\]
where \(E\) is the number of capability-phase epochs and \(B\) is the mini-batch size. As in learning-rate scheduling, setting \(k\) too small may under-explore high-impact capability improvements, while setting it too large may delay lower-risk behavioral refinement. In our experiments, a single capability-phase epoch (\(E=1\)) works well, allowing the search to traverse the training set once with high-impact patches before switching to targeted steering adjustments.

\paragraph{Takeaway.}
\proposed{} can be viewed as mini-batch offline learning over externalized agent parameters. Its design differs from standard gradient training because harness parameters are textual and non-differentiable; the performance signal is sparse and delayed; and credit assignment across long-horizon traces is ambiguous. The additional machinery is therefore not incidental. Trace collection, diagnosis, patch-as-intervention, same-batch verification, development-set validation, reflection, EvoDAG memory, and evolution are the symbolic counterparts of the mechanisms that are compactly handled by loss functions, backpropagation, optimizer state, and checkpointing in traditional mini-batch training.

\section{Additional Details on Experiment Setup}
\label{app:more_details_exp_setup}

\paragraph{Implementation of EvoDAG.}
\label{para:evo_dag_cli}
The DAG underlying EvoDAG can be implemented straightforwardly, but providing effective agent access to the DAG is more challenging: as candidates and traces accumulate across iterations, navigating EvoDAG directly becomes impractical. Rather than serializing the full DAG into the prompt, we introduce the \texttt{evo-dag} command-line interface (CLI) to facilitate interaction between the CA-SDK and EvoDAG. The interface provides on-demand access to the DAG’s structured summaries, including patch history, lessons learned, the scenario registry, and harness code diffs. This allows the agent to identify relevant context before drilling down into specific raw traces or source files for detailed analysis. The full command set is summarized in~\autoref{tab:evo_dag_commands}.
 
\begin{table}[h]
\centering
\footnotesize
\caption{Commands exposed by the \texttt{evo-dag} CLI, grouped into read operations for scoped EvoDAG views and write operations for per-session node or scenario updates.}
\label{tab:evo_dag_commands}
\small
\begin{tabular}{@{}p{0.32\linewidth} p{0.32\linewidth} p{0.30\linewidth}@{}}
\toprule
\textbf{Command} & \textbf{Purpose} & \textbf{When to use} \\
\midrule
\multicolumn{3}{@{}l}{\textit{Read operations}} \\
\midrule
\texttt{evo-dag summary}
& DAG topology, best candidate, all edges
& Start of session; quick orientation \\
\addlinespace
\texttt{evo-dag show history}
& Full patch history: diffs, reflections, lessons
& Deep dive into prior attempts \\
\addlinespace
\texttt{evo-dag show node <idx>}
& Node details: scores, intent, verdict, output dirs
& Inspecting a specific candidate \\
\addlinespace
\texttt{evo-dag show edge <parent> <child>}
& Code diff, per-scenario impacts, files changed
& Understanding a specific patch \\
\addlinespace
\texttt{evo-dag show scenario <id>}
& Per-scenario history, root causes, attempted fixes
& Before diagnosing a failing scenario \\
\addlinespace
\texttt{evo-dag show current-batch}
& Current mini-batch scenario IDs and output dirs
& Orienting to the current iteration \\
\addlinespace
\texttt{evo-dag show lineage}
& DAG lineage with edge types
& Understanding branching structure \\
\addlinespace
\texttt{evo-dag show lessons}
& Accumulated good/bad patterns
& Checking known patterns before patching \\
\midrule
\multicolumn{3}{@{}l}{\textit{Write operations}} \\
\midrule
\texttt{evo-dag update-selection}
& Record selected parent candidate(s) and the reasoning behind the choice
& Evolution Session, after parent selection \\
\addlinespace
\texttt{evo-dag update-intent}
& Record target scenarios, diagnosis, approach, files changed, and change summary
& Diagnosis-Patch Session, after applying patches \\
\addlinespace
\texttt{evo-dag update-reflection}
& Record per-scenario status (\texttt{fixed} / \texttt{regressed} / \texttt{still\_failing} / \texttt{still\_passing}), root cause, post-patch explanation, prevention or next step, and generalization note
& Reflection Session, per scenario \\
\bottomrule
\end{tabular}
\end{table}

\paragraph{Adapting Baselines to Our Benchmarks.}
GEPA treats the system prompt as a single unified string. By contrast, the default ReAct-based agent in GAIA2 distributes its system prompt across multiple sections, such as core behavioral principles, the ReAct JSON tool-calling format, and Meta's Agents Research Environments (ARE) simulation-environment instructions. These sections are defined as separate variables and concatenated only at inference time, creating a structural mismatch that prevents the direct application of GEPA. To address this mismatch, we concatenate the sections into a single string, using \texttt{=====~VAR\_NAME~=====} delimiters to mark section boundaries, and instruct GEPA's reflection LLM to preserve these delimiters during prompt evolution. The evolved prompt is then split along the delimiters, and the resulting sections are passed to CA-SDK to update the corresponding variables.
For SBP and TB2, we adopt SWE-agent and Terminus 2 as the base harnesses, respectively; since both agents define their system prompts as a single unified string, GEPA can be applied directly without further modification. 

The original Meta-Harness assumes that the harness is contained within a single Python script and requires inheritance from a fixed base agent class. In the GAIA2 default agent, however, the harness may be distributed across an entire repository. We therefore extend the Meta-Harness adaptation to the repository level by instructing CA-SDK to apply the same patching procedure used in the original Meta-Harness across the full repository. This enables holistic harness optimization beyond the single-file setting.

\paragraph{Benchmark-Specific Evaluation Protocol.}
For each benchmark, we follow its official evaluation infrastructure to ensure that the reported results are directly comparable to prior leaderboard numbers.

\textit{GAIA2.} We run the default ReAct-based agent through ARE using the \texttt{are-run} command, which executes each task in its corresponding sandboxed \texttt{Universe} and records the full trajectory. Following the GAIA2 protocol~\cite{GAIA2}, we use \texttt{Llama-3.3-70B-Instruct} as the judge model to determine task success from the final agent state.

\textit{SWE-Bench Pro.} Following the official protocol of~\cite{SWE-Bench-Pro}, each agent-generated patch is applied inside a per-instance Docker image preconfigured with the repository at the issue commit. The patch is then verified using the instance's \emph{fail-to-pass} and \emph{pass-to-pass} unit tests. A task is counted as resolved only if all fail-to-pass tests pass and no pass-to-pass test regresses.

\textit{Terminal-Bench 2.0.} Following~\cite{TerminalBench2}, we use \texttt{Harbor} to run each task in an isolated Docker container. Success is determined by a task-specific test script that inspects the final terminal and system state.

\paragraph{Data Splits across Benchmarks.}
To evaluate whether optimized harnesses generalize beyond the tasks seen during optimization, we construct train, development, and test splits separately for each benchmark, as summarized in~\autoref{tab:data_splits}. For benchmarks with a natural grouping structure, we split by task group rather than by individual tasks. In GAIA2, the split axis is the \texttt{Universe}, where each \texttt{Universe} corresponds to a distinct persona and simulated digital environment. This design ensures that the test set contains tasks from groups unseen during optimization, providing a stronger measure of out-of-distribution generalization than random task-level splits. In SWE-Bench Pro, the split axis is the repository, which also induces shifts in programming language, codebase structure, and issue distribution. This design ensures that the test set contains tasks from groups unseen during optimization, providing a stronger measure of out-of-distribution generalization than random task-level splits. For Terminal-Bench 2.0, tasks span diverse domains but do not provide a natural grouping axis; we therefore use a uniform random partition into train, dev, and test sets. 

\begin{table}[htbp]
\centering
\footnotesize
\setlength{\tabcolsep}{5pt}
\renewcommand{\arraystretch}{1.1}
\caption{Data splits across the three benchmarks. To evaluate generalization 
under distribution shift, we partition each benchmark such that train, 
development, and test sets contain tasks from \emph{disjoint task groups}---
\texttt{Universes} (personas) for GAIA2 and repositories (programming 
languages) for SWE-Bench Pro---rather than random task-level splits. Terminal-Bench 2.0 contains only 89 tasks across diverse 
domains and offers no natural grouping axis, so we adopt a uniform random 
partition.}
\label{tab:data_splits}
\begin{tabular}{@{}llllr@{}}
\toprule
\textbf{Benchmark} & \textbf{Split Axis} & \textbf{Split} & \textbf{Task Group} & \textbf{\# Tasks} \\
\midrule
\multirow{5}{*}{\textbf{GAIA2}}
  & \multirow{5}{*}{\texttt{Universe} (persona)}
  & Train & \texttt{Universe 29}            & 75  \\
\cmidrule(l){3-5}
  & & Dev   & \texttt{Universe 30}            & 65  \\
\cmidrule(l){3-5}
  & & \multirow{3}{*}{Test}
          & \texttt{Universe 21}            & 107 \\
  & & & \texttt{Universe 22}                & 112 \\
  & & & \texttt{Universe 27}                & 81  \\
\midrule
\multirow{6}{*}{\textbf{SWE-Bench Pro}}
  & \multirow{6}{*}{Repository (language)}
  & Train & \texttt{qutebrowser} (Python)   & 79 \\
\cmidrule(l){3-5}
  & & \multirow[t]{2}{*}{Dev}
          & \texttt{Vuls} (Go)              & 40 \\
  & & & \texttt{NodeBB} (JavaScript)        & 40 \\
\cmidrule(l){3-5}
  & & \multirow[t]{3}{*}{Test}
          & \texttt{Ansible} (Python)       & 96 \\
  & & & \texttt{Flipt} (Go)                 & 85 \\
  & & & \texttt{Element-web} (TypeScript)   & 56 \\
\midrule
\multirow{3}{*}{\textbf{Terminal-Bench 2.0}}
  & \multirow{3}{*}{Random$^{\dagger}$}
  & Train & ---  & 30 \\
\cmidrule(l){3-5}
  & & Dev   & ---  & 19 \\
\cmidrule(l){3-5}
  & & Test  & ---  & 40 \\
\bottomrule
\end{tabular}
\par\smallskip
\raggedright\scriptsize
$^{\dagger}$\,Only 89 tasks across heterogeneous domains (system administration, machine learning, cybersecurity); no natural axis for distribution-shifted splits exists.
\end{table}

\paragraph{Optimization Budgets.}
For \proposed{} and GEPA, we optimize for 2 epochs on GAIA2 and SWE-Bench-Pro, and for 4 epochs on Terminal-Bench 2.0; the doubled budget on TB2 compensates for its training set being roughly half the size of GAIA2.
For Meta-Harness, applying the same epoch-based budget would be misleading: unlike \proposed{} and GEPA, which perform mini-batch optimization, Meta-Harness operates in a full-batch regime, so a 2-epoch budget corresponds to only two patch updates and would unfairly handicap the baseline. 
To ensure a fair comparison, we instead match optimization budgets in terms of total task executions consumed during training. 
Concretely, we train Meta-Harness for 20 epochs on GAIA2 and 15 epochs on Terminal-Bench 2.0, and 8 epochs on SWE-Bench-Pro, ensuring its total task executions are no smaller than those used by \proposed{} on every benchmark.

\section{Robustness to Optimization Stochasticity}
\label{app:optimization_stochasticity}

\begin{table}[!h]
\centering
\small
\setlength{\tabcolsep}{5pt}
\caption{\textbf{Robustness to optimization stochasticity}. Pass@1 on GAIA2
Universe~22 for harnesses obtained from two independent optimization runs.
Test performance is reported as mean $\pm$ standard deviation over three
executions.}
\label{tab:optimization_stochasticity}
\sisetup{
separate-uncertainty = true,
table-format         = 2.1(2),
detect-weight        = true,
retain-zero-uncertainty = true,
detect-inline-weight = math,
}
\begin{tabular}{lS}
\toprule
\textbf{Method} &
{\textbf{GAIA2 Universe 22 (112 scenarios; Pass@1)}} \\
\midrule
Default Agent (Manual)        & 51.5(49) \\
GEPA (Run 1)                  & 47.9(34) \\
GEPA (Run 2)                  & 50.6(05) \\
Meta-Harness (Run 1)          & 51.5(52) \\
Meta-Harness (Run 2)          & 51.2(22) \\
\proposed{} (Run 1)           & \bfseries 60.7(24) \\
\proposed{} (Run 2)           & 58.6(05) \\
\bottomrule
\end{tabular}
\end{table}

Our main experiments perform one optimization run for each method because
end-to-end harness optimization is substantially more expensive than repeated
test-time evaluation. To assess whether the observed gains are sensitive to
stochastic optimization trajectories, we conduct an additional independent
optimization run for \proposed{}, GEPA, and Meta-Harness on GAIA2.
We evaluate each resulting harness on Universe~22, the largest held-out test
universe with 112 scenarios, using three repeated executions per harness.

\autoref{tab:optimization_stochasticity} reports the results.
The second \proposed{} optimization run achieves 58.6\% Pass@1, only 2.1
percentage points below the first run and 7.1 points above the default agent.
It also outperforms the independently rerun GEPA and Meta-Harness harnesses by
8.0 and 7.4 percentage points, respectively.
Although two optimization trajectories are insufficient for a comprehensive
statistical characterization of the optimizer, the consistent gains across
independent runs indicate that the improvement of \proposed{} is not specific
to a single favorable search trajectory.

\section{Robustness to Training-Distribution Shift}
\label{app:training_distribution_robustness}

\begin{table}[!h]
\centering
\small
\setlength{\tabcolsep}{5pt}
\caption{\textbf{Robustness to training-distribution shift}. Pass@1 on GAIA2
Universe~22 when \proposed{} is independently optimized using different
training universes. Values are mean $\pm$ standard deviation over three
test executions.}
\label{tab:training_distribution_robustness}
\sisetup{
separate-uncertainty = true,
table-format         = 2.1(2),
detect-weight        = true,
retain-zero-uncertainty = true,
detect-inline-weight = math,
}
\begin{tabular}{lS}
\toprule
\textbf{Method} &
{\textbf{GAIA2 Universe 22 (112 scenarios; Pass@1)}} \\
\midrule
Default Agent (Manual)                     & 51.5(49) \\
\proposed{} (Run 1; Universe 29)            & \bfseries 60.7(24) \\
\proposed{} (Run 2; Universe 29)            & 58.6(05) \\
\proposed{} (Different Train Set; Universe 24)
                                            & 57.4(21) \\
\bottomrule
\end{tabular}
\end{table}

The main GAIA2 experiments optimize the harness using Universe~29 as the
training set. To examine whether the resulting gains depend critically on this
particular training distribution, we replace Universe~29 with the comparably
sized Universe~24 and independently optimize \proposed{} while keeping the
remaining optimization and evaluation protocol unchanged.
The resulting harness is evaluated three times on Universe~22.

As shown in \autoref{tab:training_distribution_robustness}, the harness
optimized on Universe~24 achieves 57.4\% Pass@1.
Despite being optimized using a different training universe, it improves over
the default agent by 5.9 percentage points and remains within 1.2 points of
the second independent \proposed{} run trained on Universe~29.
These results suggest that \proposed{} can discover effective harness updates
from different persona and task distributions, rather than relying on
properties specific to the original Universe~29 training split.

\section{Cross-Model Transferability to Weaker Agent Backbone}
\label{app:appendix_cross_model}

While \autoref{tab:eval_results_gaia2} reports results with \texttt{Claude Opus~4.6} serving as both the optimizer and the agent backbone,
a key practical question is whether harnesses optimized by a strong model retain their benefits when deployed with a weaker task agent.
To test this, we re-evaluate all methods on GAIA2 using \texttt{Claude Haiku~4.5} as the task agent backbone while keeping the harnesses unchanged from those produced during Opus-based optimization. \autoref{tab:eval_results_gaia2_haiku} shows that \proposed{} achieves an overall improvement of $+5.6$\,pp over the default agent, demonstrating effective cross-model transferability. 
Notably, \proposed{} consistently outperforms all baseline methods and ablation settings across all evaluated universes.

\begin{table}[!h]
\centering
\small
\setlength{\tabcolsep}{4pt}
\caption{\textbf{Cross-model transferability}. Pass@1 success rates on the test-set of GAIA2 using
\texttt{Claude Haiku 4.5} as the agent backbone, with harnesses optimized by
\texttt{Claude Opus 4.6}.
Values are mean $\pm$ standard deviation over three runs.
}
\label{tab:eval_results_gaia2_haiku}
\sisetup{
  separate-uncertainty = true,
  table-format         = 2.1(2),
  detect-weight        = true,
  retain-zero-uncertainty = true,
  detect-inline-weight = math,
}
\begin{tabular}{lSSSS}
\toprule
\multicolumn{1}{c}{\multirow{2}{*}{\textbf{Harness (Type)}}}
 & \multicolumn{3}{c}{\textbf{GAIA2 Universe (Pass@1)}}
 & \multicolumn{1}{c}{\multirow{2}{*}{\textbf{Avg.}}} \\
\cmidrule(lr){2-4}
 & {\textbf{21 (107)}} & {\textbf{22 (112)}} & {\textbf{27 (81)}} & \\
\midrule
Default Agent (Manual)                      & 31.2(59)           & 24.4(14)           & 36.2(19)           & 30.0(18)          \\
GEPA (Auto)                                 & 33.6(19)           & 24.4(05)           & 38.3(12)           & 31.4(04)           \\
Meta-Harness (Auto)                         & 30.8(00)           & 26.8(47)           & 34.2(40)           & 30.2(27)           \\
\textbf{\proposed{} (Auto)}                 & \bfseries 38.3(00) & \bfseries 30.4(18) & \bfseries 39.1(26) & \bfseries 35.6(08) \\
\midrule
w/o In-depth Diagnosis (Auto)               & 36.1(55)           & 19.0(52)           & 34.2(07)           & 29.2(34)           \\
w/o Structured Intervention (Auto)          & 32.1(30)           & 28.0(40)           & 37.4(58)           & 32.0(35)           \\
w/o Generalization-aware Selection (Auto)   & 28.3(29)           & 23.2(41)           & 30.0(38)           & 26.9(13)           \\
\bottomrule
\end{tabular}

\smallskip
\end{table}

\section{Fine-Grained Ablations of Structured Intervention and
Generalization-Aware Selection}
\label{app:fine_grained_ablations}

\begin{table}[!h]
\centering
\small
\setlength{\tabcolsep}{5pt}
\caption{\textbf{Fine-grained ablations}. Pass@1 on GAIA2 Universe~22.
Values are mean $\pm$ standard deviation over three runs.}
\label{tab:fine_grained_ablations}
\sisetup{
separate-uncertainty = true,
table-format         = 2.1(2),
detect-weight        = true,
retain-zero-uncertainty = true,
detect-inline-weight = math,
}
\begin{tabular}{lS}
\toprule
\textbf{Method} &
{\textbf{GAIA2 Universe 22 (112 scenarios; Pass@1)}} \\
\midrule
Default Agent (Manual)                         & 51.5(49) \\
\proposed{}                                    & \bfseries 60.7(24) \\
\midrule
w/o Structured Intervention                    & 53.3(76) \\
w/o Phase Scheduling                           & 54.8(44) \\
\midrule
w/o Generalization-Aware Selection             & 44.9(67) \\
w/o Dev-Set Filtering                          & 50.0(50) \\
\bottomrule
\end{tabular}
\end{table}

The main ablations remove each design principle of \proposed{} as a whole.
We further disentangle the individual mechanisms underlying
\emph{structured intervention} and \emph{generalization-aware selection}.
For structured intervention, the main \emph{w/o Structured Intervention}
ablation removes both the patch taxonomy and Phased Patch Scheduling.
We therefore introduce a finer-grained \emph{w/o Phase Scheduling} variant
that retains the patch taxonomy while removing only the phased schedule.
Similarly, the main \emph{w/o Generalization-Aware Selection} ablation removes
both development-set filtering and Reflection with EvoDAG.
We introduce a \emph{w/o Dev-Set Filtering} variant that retains Reflection
with EvoDAG but removes development-set filtering.
For each variant, we rerun the full optimization procedure and evaluate the
resulting harness on GAIA2 Universe~22 using three repeated test executions.

\autoref{tab:fine_grained_ablations} shows the results.
Removing only Phased Patch Scheduling reduces Pass@1 from 60.7\% to 54.8\%,
a drop of 5.9 percentage points.
Removing the patch taxonomy in addition further reduces performance to
53.3\%, indicating an additional 1.5-point contribution from structuring the
patch space.
For generalization-aware selection, removing development-set filtering reduces
Pass@1 from 60.7\% to 50.0\%, while additionally removing Reflection with
EvoDAG further reduces performance to 44.9\%.
Thus, development-set filtering accounts for the larger effect in this
comparison, while Reflection with EvoDAG provides an additional 5.1-point
gain.
Together, these finer-grained ablations indicate that both mechanisms within
each design principle contribute to final harness performance.

\section{Patch Durability Analysis}
\label{app:patch_durability}

A desirable harness update should not only repair the scenarios that motivate
the patch, but should also preserve behavior that was already correct.
We therefore analyze patch durability by comparing task outcomes before and
after each generated patch on the same training mini-batch.
We define the \emph{fix rate} as the fraction of previously failing scenarios
that pass after applying the patch, and the \emph{regression rate} as the
fraction of previously passing scenarios that fail after the patch.

We first group patches according to the three harness components used by the
patch taxonomy: Prompt, Tool, and Middleware.
As shown in \autoref{tab:patch_durability_category}, the three categories
exhibit similar fix rates of 57--59\%, while Tool patches have a moderately
higher regression rate of 19\%, compared with 14\% for Prompt and Middleware
patches.
Thus, the component being edited alone does not reveal a clear durability
pattern.

\begin{table}[!h]
\centering
\small
\caption{\textbf{Patch durability by harness component}. Fix and regression
rates are measured by comparing pre- and post-patch outcomes on the same
training mini-batch.}
\label{tab:patch_durability_category}
\begin{tabular}{lcc}
\toprule
\textbf{Patch Category} & \textbf{Fix Rate (\%)} &
\textbf{Regression Rate (\%)} \\
\midrule
Prompt      & 59 & 14 \\
Tool        & 57 & 19 \\
Middleware  & 57 & 14 \\
\bottomrule
\end{tabular}
\end{table}

One reason for the weak separation across these categories is that they mix
different intervention mechanisms.
For example, Tool patches include executable changes such as new tools and
implementation fixes, but also textual changes to tool descriptions.
Similarly, Middleware patches include both executable infrastructure or
agent-loop changes and text-only PreToolUse hooks.
We therefore reclassify all generated patches using the higher-level
Capability--Steering taxonomy from the main paper.
Capability Patches modify executable functionality or orchestration logic,
whereas Steering Patches alter textual instructions without changing the
underlying executable capability.

\autoref{tab:patch_durability_mechanism} reveals a clearer difference.
Steering Patches achieve a slightly higher fix rate than Capability Patches
(58\% vs.\ 55\%), but their regression rate is more than twice as high
(17\% vs.\ 8\%).
Capability Patches therefore obtain a comparable rate of local repairs while
reducing regressions by 9 percentage points.
This result suggests that capability-level interventions tend to produce more
durable updates, whereas textual steering is more susceptible to spilling over
to scenarios outside the intended scope.
This quantitative pattern is also consistent with our qualitative examples:
well-scoped executable fixes often address deterministic capability
limitations, while overly broad hooks or prompt rules can affect unrelated
scenarios.

\begin{table}[!h]
\centering
\small
\caption{\textbf{Patch durability by intervention mechanism}. Capability and
Steering Patches exhibit similar fix rates, but Steering Patches induce
substantially more regressions.}
\label{tab:patch_durability_mechanism}
\begin{tabular}{lcc}
\toprule
\textbf{Patch Type} & \textbf{Fix Rate (\%)} &
\textbf{Regression Rate (\%)} \\
\midrule
Steering Patch   & 58 & 17 \\
Capability Patch & 55 & 8  \\
\bottomrule
\end{tabular}
\end{table}

\section{Optimization Efficiency on Terminal-Bench 2.0}
\label{appendix:optimization_efficiency_other_benchmarks}

To assess whether \proposed{}'s optimization efficiency generalizes beyond
GAIA2, we report compute and learning efficiency on Terminal-Bench~2.0,
which targets a distinct agentic capability: long-horizon
command-line problem solving. We follow the same protocol as in our main
GAIA2 experiments (Section~\ref{sec:experiments}), comparing \proposed{}
against Meta-Harness and GEPA.

\begin{figure}[!h]
\centering
    \begin{subfigure}[b]{0.49\linewidth}
        \centering
        \includegraphics[width=\linewidth]{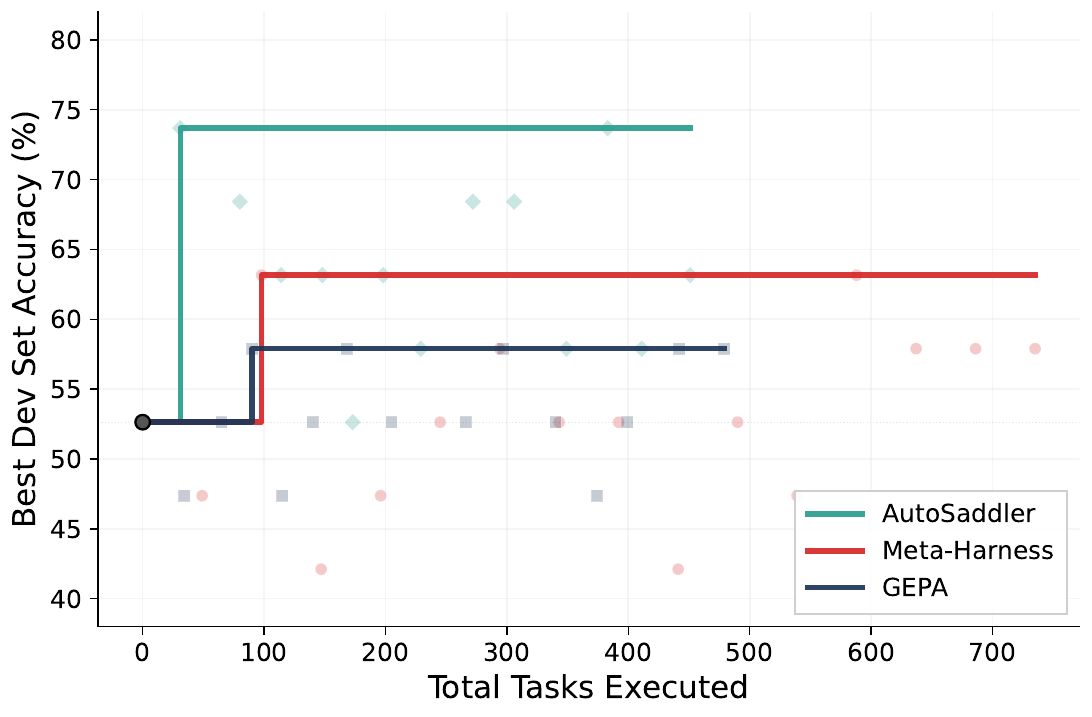}
        \caption{Compute Efficiency}
        \label{fig:tbench2_compute_efficiency}
    \end{subfigure}
    \hfill
    \begin{subfigure}[b]{0.49\linewidth}
        \centering
        \includegraphics[width=\linewidth]{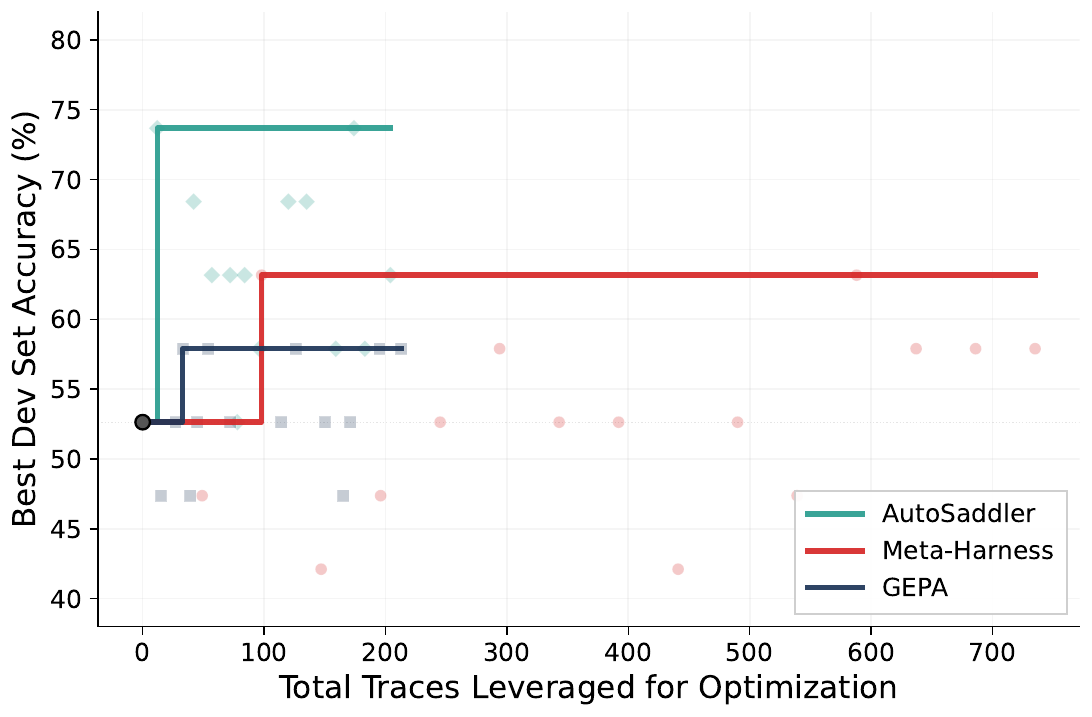}
        \caption{Learning Efficiency}
        \label{fig:tbench2_learning_efficiency}
    \end{subfigure}
    \caption{Comparison of optimization performance and efficiency on Terminal-Bench~2.0. \textbf{(a)} \proposed{} reaches 73.7\% dev accuracy with only 31 task executions, whereas Meta-Harness requires 98 executions to plateau at 63.2\% and GEPA reaches 57.9\% at 90 executions. \textbf{(b)} When measured by the number of execution traces leveraged for optimization, \proposed{} achieves its best performance after leveraging only 12 traces, over 8$\times$ fewer than Meta-Harness (98 traces), while surpassing it by 10.5 percentage points.}
    \label{fig:tbench2_optimization_comparison}
\end{figure}

\autoref{fig:tbench2_optimization_comparison} reports compute and learning
efficiency on Terminal-Bench~2.0. From the common 52.6\% starting point,
\proposed{} reaches \textbf{73.7\%} dev accuracy after only 31 task executions
and 12 leveraged traces, outperforming Meta-Harness (63.2\%) by 10.5 percentage
points and GEPA (57.9\%) by 15.8 percentage points. The ${\sim}$8$\times$
reduction in leveraged traces relative to Meta-Harness (98 traces) mirrors the
trend observed on GAIA2, indicating that \proposed{}'s diagnosis-guided patching
delivers consistent optimization efficiency gains on terminal-centric tasks
where failures are dominated by tool-use and environment-interaction errors
rather than multi-hop reasoning.

\section{End-to-End Optimization Cost Characterization}
\label{app:full_optimization_cost}

Task-agent rollouts alone do not fully characterize the computational cost of
automatic harness optimization.
In addition to harness evaluation, each optimizer may incur substantial
optimizer-side LLM overhead for candidate generation, diagnosis, patching,
reflection, and candidate selection.
We therefore separately profile (i) optimizer-side LLM overhead and
(ii) task-agent evaluation cost on GAIA2.

\paragraph{Optimizer-side cost.}
\autoref{tab:optimizer_side_cost} reports the number of generated, rejected,
and accepted patches together with the average wall-clock time, monetary cost,
LLM calls, and token usage per generated patch.
GEPA incurs the lowest optimizer-side cost at \$5.50 per patch, but searches
only over the system prompt.
Meta-Harness and \proposed{} instead optimize broader harness components and
therefore provide a more direct comparison.
\proposed{} costs \$14.56 per generated patch, \$1.91 more than Meta-Harness,
while requiring 533 seconds rather than 883 seconds per patch, corresponding
to 39.6\% lower wall-clock time.
This difference holds despite \proposed{} explicitly performing diagnosis,
structured patch generation, reflection, and evolution.

\begin{table}[!h]
\centering
\scriptsize
\setlength{\tabcolsep}{3pt}
\caption{\textbf{Optimizer-side cost on GAIA2}. Runtime, monetary cost, LLM
calls, and token usage are averaged per generated patch.}
\label{tab:optimizer_side_cost}

\resizebox{\textwidth}{!}{
\begin{tabular}{lrrrrrrrrr}
\toprule
\textbf{Method}
& \shortstack{\textbf{Generated}\\\textbf{Patches}}
& \shortstack{\textbf{Rejected}\\\textbf{Patches}}
& \shortstack{\textbf{Accepted}\\\textbf{Patches}}
& \shortstack{\textbf{Wall-Clock}\\\textbf{Time (s) / Patch}}
& \shortstack{\textbf{Cost (\$)}\\\textbf{/ Patch}}
& \shortstack{\textbf{LLM Calls}\\\textbf{/ Patch}}
& \shortstack{\textbf{Output Tokens}\\\textbf{/ Patch}}
& \shortstack{\textbf{Cache-Creation}\\\textbf{Input Tokens / Patch}}
& \shortstack{\textbf{Cache-Read}\\\textbf{Input Tokens / Patch}} \\
\midrule

GEPA
& 76 & 58 & 18
& 386 & 5.50 & 21.3
& 11,416 & 156,788 & 280,940 \\

Meta-Harness\tablefootnote{Meta-Harness does not employ an explicit
patch-acceptance gate; therefore, rejected and accepted patch counts are
not applicable.}
& 20 & -- & --
& 883 & 12.65 & 66.9
& 43,429 & 250,466 & 3,528,484 \\

\proposed{}
& 39 & 19 & 20
& 533 & 14.56 & 76.1
& 21,464 & 430,933 & 3,246,158 \\

\bottomrule
\end{tabular}
}
\end{table}

\paragraph{Task-agent evaluation cost.}
Harness evaluation is considerably more expensive than the optimizer-side
operations above.
A single GAIA2 task-agent rollout requires, on average, 20.2 LLM calls,
550,988 input tokens, 7,380 output tokens, and 203.9 seconds of wall-clock
time, as summarized in \autoref{tab:task_agent_rollout_cost}.

\begin{table}[!h]
\centering
\small
\caption{\textbf{Average task-agent evaluation cost per GAIA2 rollout}.}
\label{tab:task_agent_rollout_cost}
\begin{tabular}{lr}
\toprule
\textbf{Metric} & \textbf{Average per Rollout} \\
\midrule
LLM calls       & 20.2 \\
Input tokens    & 550,988 \\
Output tokens   & 7,380 \\
Wall-clock time & 203.9 s \\
\bottomrule
\end{tabular}
\end{table}

The evaluation protocols of the methods differ substantially in how often
this expensive operation is invoked.
Meta-Harness evaluates the union of all 140 GAIA2 training and development
scenarios for every candidate update.
In contrast, \proposed{} evaluates six training scenarios per optimization
iteration and invokes the 65-scenario development-set evaluation only for
patches that improve the training mini-batch.
Consequently, \proposed{} reaches 67.7\% development accuracy after 391
task-agent rollouts, already exceeding the 61.5\% peak reached by Meta-Harness
after 1,400 rollouts.
Thus, while \proposed{} incurs moderately higher optimizer-side monetary cost
per generated patch, this overhead is offset by substantially more selective
use of expensive task-agent evaluations and higher performance under a much
smaller rollout budget.

\section{Search Trajectory Visualization}
\label{app:search_trajectory_visualization}
\begin{figure}[!h]
\centering
    \includegraphics[width=0.9\linewidth]{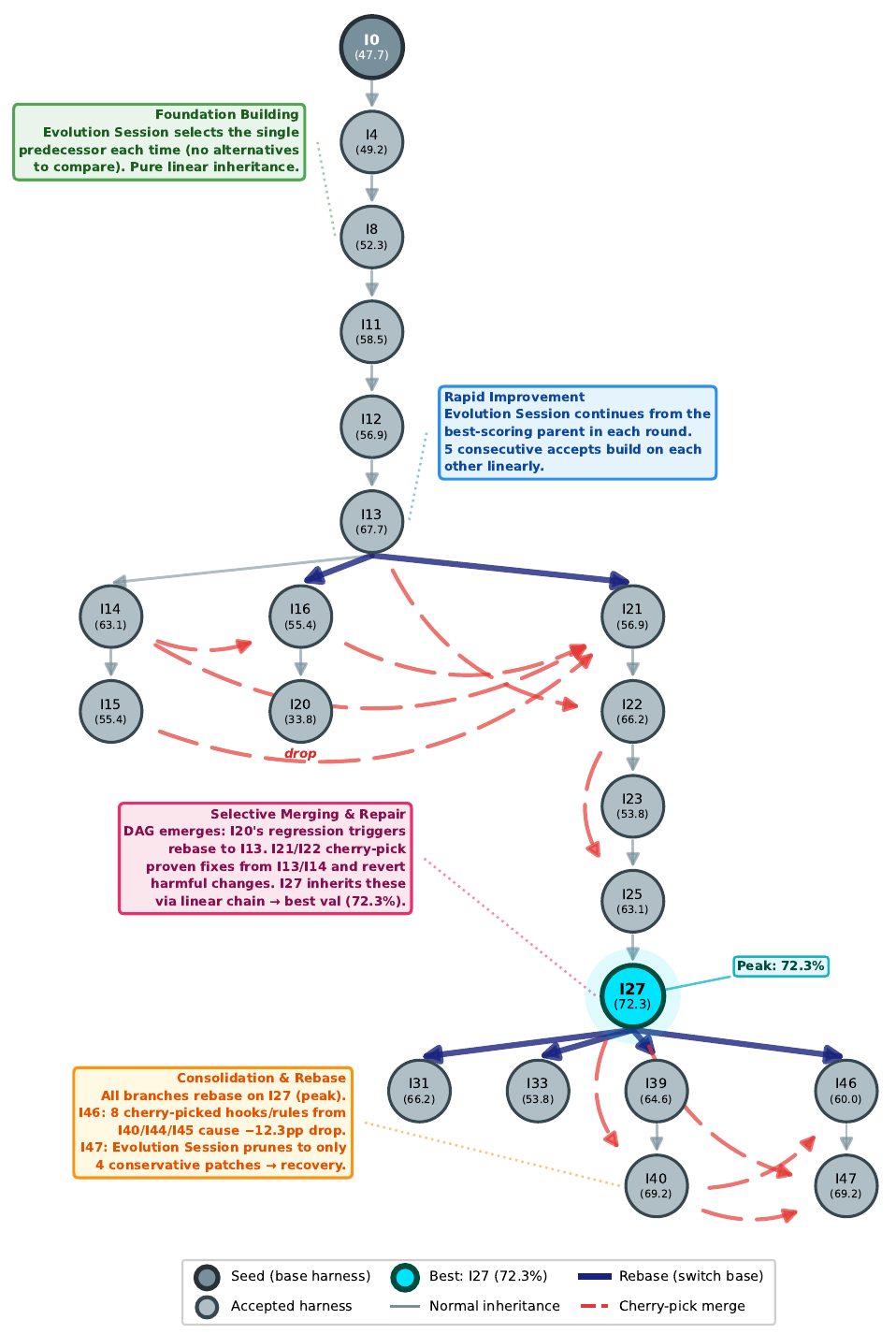}
    \caption{Evolutionary search trajectory of \proposed{} as the EvoDAG. Four phases (foundation, rapid improvement, selective merging, and consolidation) lead to a global dev-set peak of 72.3\% at Iter27, escaping the Iter20 regression and local optima. 
    }
    \label{fig:evolution_dag}
\end{figure}

To illustrate how the reflection and evolution sessions jointly shape the search trajectory, \autoref{fig:evolution_dag} visualizes the dev-score progression of \proposed{} across 50 iterations (2 epochs) on GAIA2. Each node represents a harness that was accepted on its mini-batch and subsequently evaluated on the held-out dev set (21 of 51 total candidates); the displayed score is the dev-set accuracy. Solid edges denote base inheritance (thin gray for sequential, bold dark for rebase), while red dashed edges denote cherry-pick merges, the operations that lead to a DAG structure rather than a simple chain.

The trajectory exhibits four phases. In the \emph{Foundation Building} phase (Iter0–Iter8), the evolution session selects the single available predecessor at each step, producing a linear chain. In the \emph{Rapid Improvement} phase (Iter8–Iter15), five consecutive accepts build linearly on each other, culminating at Iter13 (67.7\%). The DAG structure emerges in the \emph{Selective Merging \& Repair} phase (Iter16–Iter27): after Iter20's catastrophic regression to 33.8\% caused by a hook on a high-frequency tool, the evolution session rebases to Iter13 and cherry-picks proven fixes from Iter13 and Iter14 into Iter21 and Iter22, while reverting harmful changes identified through reflection. Iter27 inherits these curated patches via a linear chain and achieves the global peak of 72.3\%. In the \emph{Consolidation \& Rebase} phase (Iter31–Iter47), all branches rebase on Iter27. When Iter46 over-accumulates eight cherry-picked hooks and rules from Iter40/Iter44/Iter45, causing a $-$12.3 pp drop, the evolution session at Iter47 diagnoses this as collective steering overhead and prunes to only four conservative patches, recovering to 69.2\%. Overall, this trajectory highlights how \proposed{} uses reflection-guided rebasing and selective merging to preserve beneficial changes, discard harmful ones, and recover from regressions during harness search.

\section{Additional Analysis for RQ1}
\label{app:more_details_RQ1}

\paragraph{Tool-Call and File-Access Overhead of In-Depth Diagnosis}
A central design choice of \proposed{} is the in-depth diagnosis stage, which is intended to perform a more thorough investigation of failure traces than shallow reflection as in the \textit{"w/o In-Depth Diagnosis"} ablation setting. 
To verify that this stage indeed conducts substantively deeper analysis, rather than merely adding superficial overhead, we empirically measure the additional tool calls and file accesses that \proposed{} expends compared to the ablation setting. 
In \proposed{}, the Claude Agent SDK performs both diagnosis and patching, whereas in the ablated variant the SDK is responsible only for patching. The gap between the two therefore directly reflects the investigative effort attributable to in-depth diagnosis.

Counting tool calls is straightforward, as each invocation of a Claude Agent SDK tool is logged. However, accurately counting file accesses is non-trivial because the Claude Agent SDK occasionally delegates work to sub-agents, and tool calls issued inside sub-agents are not recorded in the main trace. To ensure a consistent measurement across both variants, we count file accesses for the main agent using the following criteria:
\begin{itemize}
    \item \textbf{File access through dedicated file tools.} The number of invocations of the Claude Agent SDK's dedicated file-handling tools, namely, \texttt{Read}, \texttt{Write}, \texttt{Edit}, \texttt{MultiEdit}, and \texttt{Grep}.
    \item \textbf{File access through Bash commands.} The number of executions of shell commands that read or manipulate file contents, including \texttt{cat}, \texttt{wc}, \texttt{grep}, and Python invocations that open files (e.g., \texttt{python3} or \texttt{python} scripts containing \texttt{open()} calls).
\end{itemize}
The same accounting rule is applied to both \proposed{} and the ablated variant, enabling a direct comparison of investigative effort.

\autoref{tab:diagnosis_depth} reports the average tool calls and file accesses per optimization step for \proposed{} and the \textit{"w/o In-Depth Diagnosis"} ablation setting. \proposed{} performs 69.7 tool calls and 45.5 file accesses on average, compared to 63.5 and 39.7 for the ablated variant, corresponding to 6.2 additional tool calls and 5.8 additional file accesses per step.

\begin{table}[h]
\centering
\caption{Average tool calls and file accesses per optimization step for \proposed{} and its ablated variant without in-depth diagnosis.}
\label{tab:diagnosis_depth}
\begin{tabular}{lcc}
\toprule
\textbf{Method} & \textbf{Avg.\ Tool Calls} & \textbf{Avg.\ File Accesses} \\
\midrule
w/o In-Depth Diagnosis & 63.5 & 39.7 \\
\proposed{} & 69.7 & 45.5 \\
\midrule
$\Delta$ (Attributable to In-Depth Diagnosis) & +6.2 & +5.8 \\
\bottomrule
\end{tabular}
\end{table}

\paragraph{Qualitative Case Studies.}
\begin{figure}[!h]
\centering
    \includegraphics[width=1\linewidth]{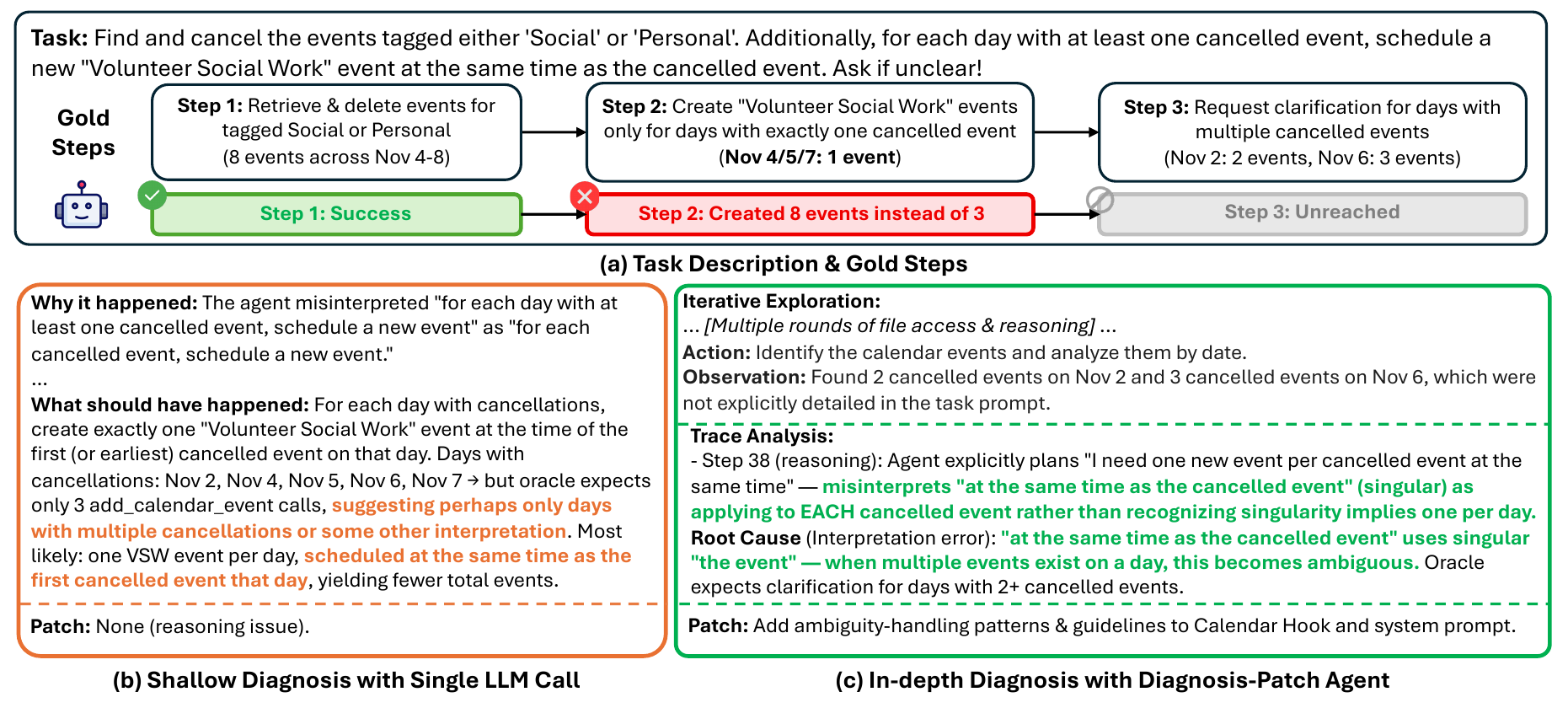}
    \caption{Case study on a calendar task with ambiguous cancellations, where AutoSaddler's in-depth diagnosis accurately identifies the true root cause via deep debugging, unlike shallow diagnosis.}    
    \label{fig:rq1_case_1} 
\end{figure}



\autoref{fig:rq1_case_1} illustrates how this depth translates into diagnostic accuracy. 
In this task, the agent should have created new events only for days with a single cancellation (Nov 4, 5, 7) and requested clarification for days with multiple cancellations (Nov 2, 6). 
Instead, ignoring task ambiguity, it created new events for all canceled events. 
Without thorough investigation, shallow diagnosis incorrectly hypothesized that the agent simply misunderstood the instruction, concluding that it should have matched the time of the first cancellation on days with multiple cancellations. 
Consequently, it misclassified the issue as a simple LLM reasoning error and bypassed patching. 
In contrast, AutoSaddler identifies the calendar events and analyzes them by date, confirming multiple cancellations on Nov 2 and 6, and cross-verifies this with the oracle events.
This enabled it to pinpoint the true root cause: the linguistic ambiguity of the singular phrase ``the event''. 
It successfully resolved the issue by patching the calendar hook and system prompt with an ambiguity-handling guideline.

\begin{figure}[!h]
\centering
    \includegraphics[width=1\linewidth]{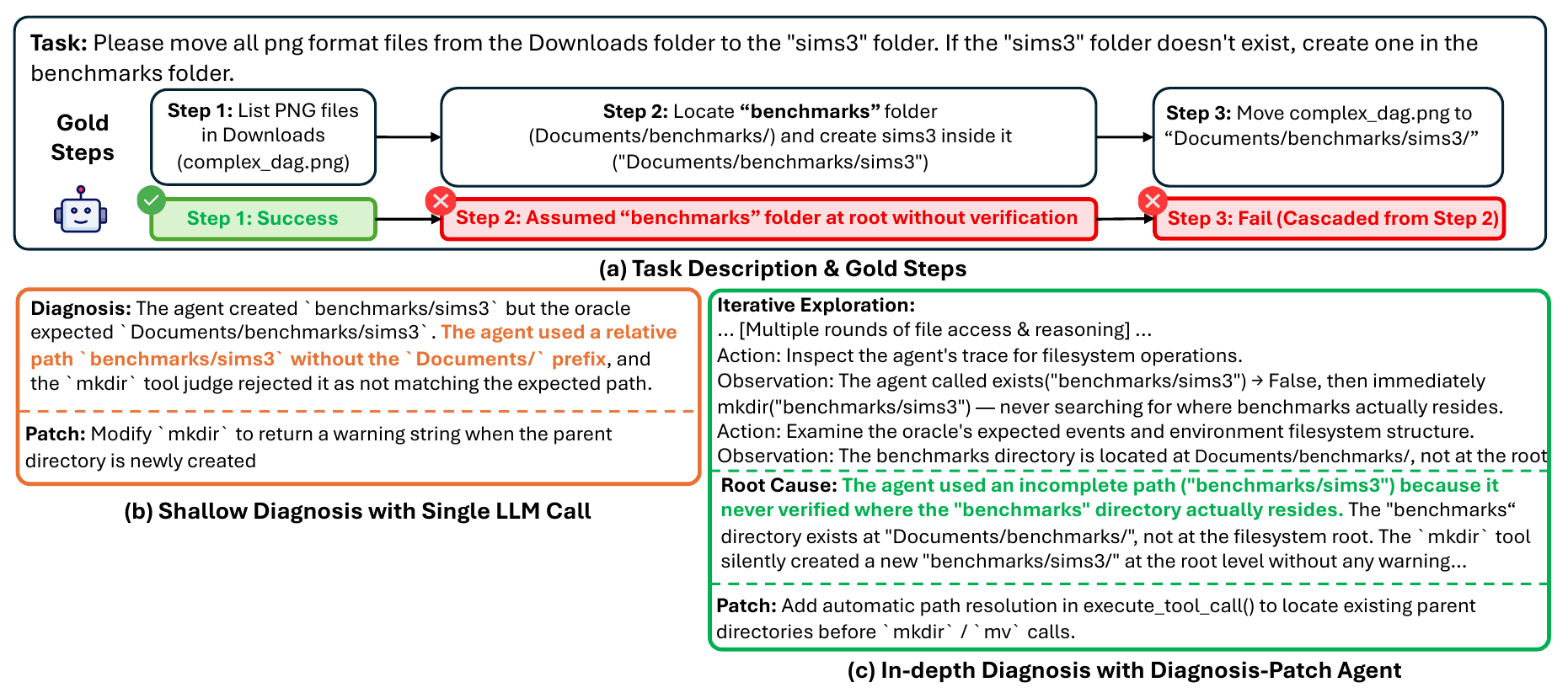}
    \caption{Case study on a file-management task that requires moving PNG files into a \texttt{sims3} folder under \texttt{benchmarks}. Shallow diagnosis misattributes the failure to a relative-path error, while AutoSaddler verifies the filesystem and locates the true cause: an unverified parent directory.}
    \label{fig:analysis_rq2_case2}
\end{figure}

\autoref{fig:analysis_rq2_case2} illustrates a complementary failure mode in which shallow reflection produces a plausible-sounding but factually incorrect diagnosis.
In this task, the agent should have located the existing \texttt{benchmarks} directory under \texttt{Documents/} before creating \texttt{sims3} inside it.
Instead, the agent created \texttt{benchmarks/sims3/} at the filesystem root without verifying where \texttt{benchmarks} actually resides.
Shallow reflection diagnoses the failure as the agent's use of an incorrect \emph{relative} path despite knowing that the benchmarks are located under \texttt{Documents/}, which is inconsistent with the agent's actual behavior of never verifying the location at all.
Consequently, it proposes a reactive patch that only modifies \texttt{mkdir} to emit a warning when the parent directory is newly created, leaving the agent's lack of pre-verification untouched.
In contrast, AutoSaddler inspects the agent's filesystem trace and observes that the agent called \texttt{exists("benchmarks/sims3")} and immediately invoked \texttt{mkdir} on the same path without searching for the true location, then cross-verifies this against the environment filesystem structure to confirm that \texttt{benchmarks} resides at \texttt{Documents/benchmarks/}.
This enabled it to pinpoint the true root cause: the absence of parent-directory verification before \texttt{mkdir} and \texttt{mv} calls.
It successfully resolved the issue by adding automatic path resolution to \texttt{execute\_tool\_call()}, eliminating the silent misplacement at its source.

\begin{figure}[!h]
\centering
    \includegraphics[width=1\linewidth]{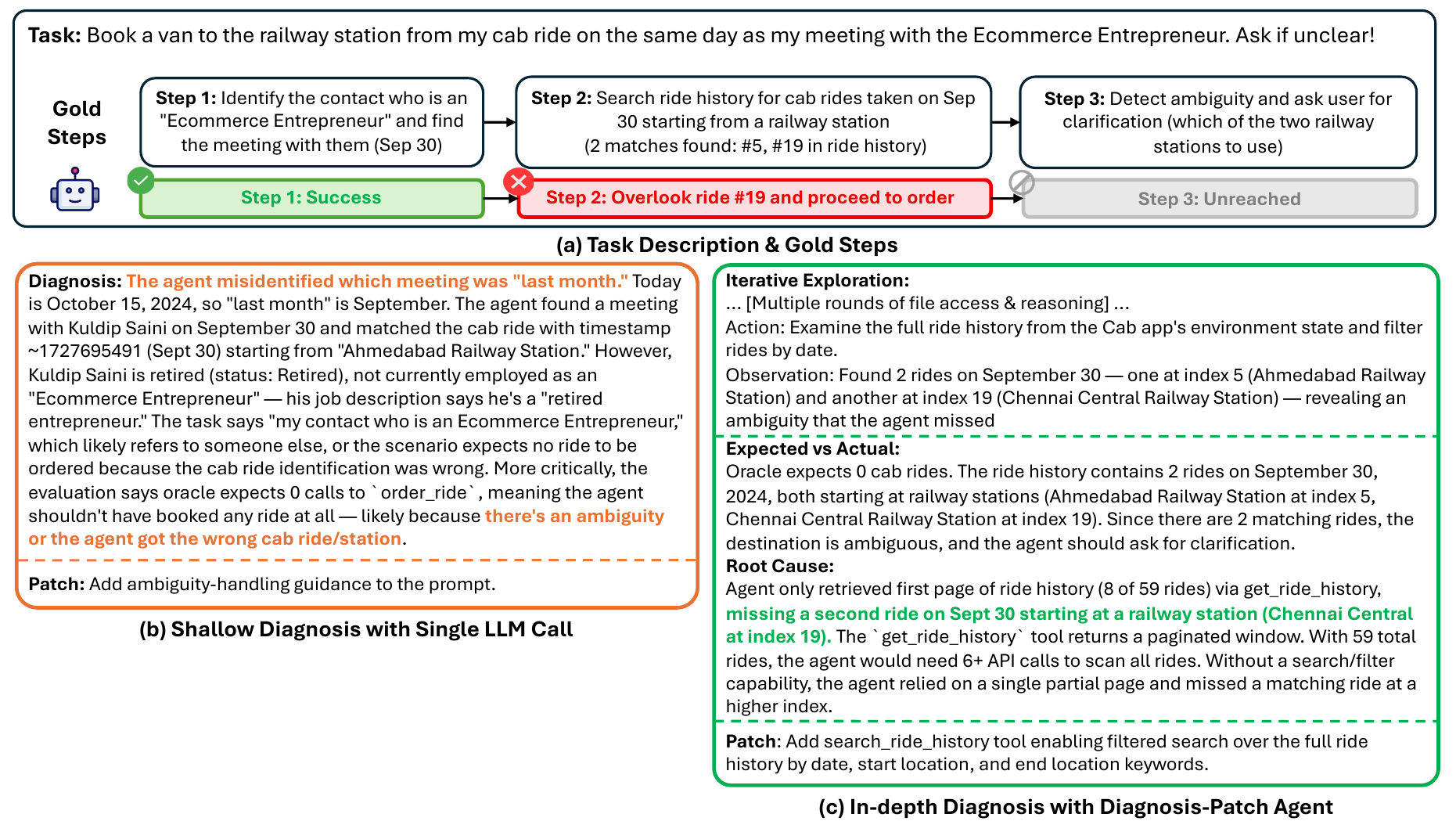}
    \caption{Case study on a cab-booking task with two candidate rides on the same day. Shallow diagnosis overlooks the second matching ride and proposes an off-target fix, while AutoSaddler inspects the full ride history and identifies the missed candidate.}
    \label{fig:analysis_rq2_case3}
\end{figure}

\autoref{fig:analysis_rq2_case3} illustrates a case in which the failure stems from a candidate that the agent never observed.
In this task, the agent should have detected two candidate rides on September 30 starting from a railway station and asked the user to disambiguate the destination.
Instead, the agent confidently booked a van using only the first matching ride, overlooking a second candidate at index~19.
Without examining the full ride history, shallow reflection fails to recognize that a second matching ride exists at index~19 and instead produces a diagnosis centered on contact identification and oracle-call counts, missing the actual cause of the failure.
Consequently, it proposes a fix that does not address the overlooked candidate.
In contrast, AutoSaddler directly examines the full ride history from the Cab app's environment state and filters by date, finding two rides on September 30, one at index~5 (Ahmedabad Railway Station) and another at index~19 (Chennai Central Railway Station), and cross-verifies this against the oracle's expected zero \texttt{order\_ride} calls.
This enabled it to pinpoint the true root cause: the agent missed the matching ride at index~19 and thus failed to detect the destination ambiguity.
It successfully resolved the issue by adding a \texttt{search\_ride\_history} tool that enables filtered search over the ride history using date, start-location, and end-location keywords, surfacing conflicting candidates in a single call and triggering the correct clarification behavior.

\begin{figure}[h]
    \centering
    \includegraphics[width=0.5\linewidth]{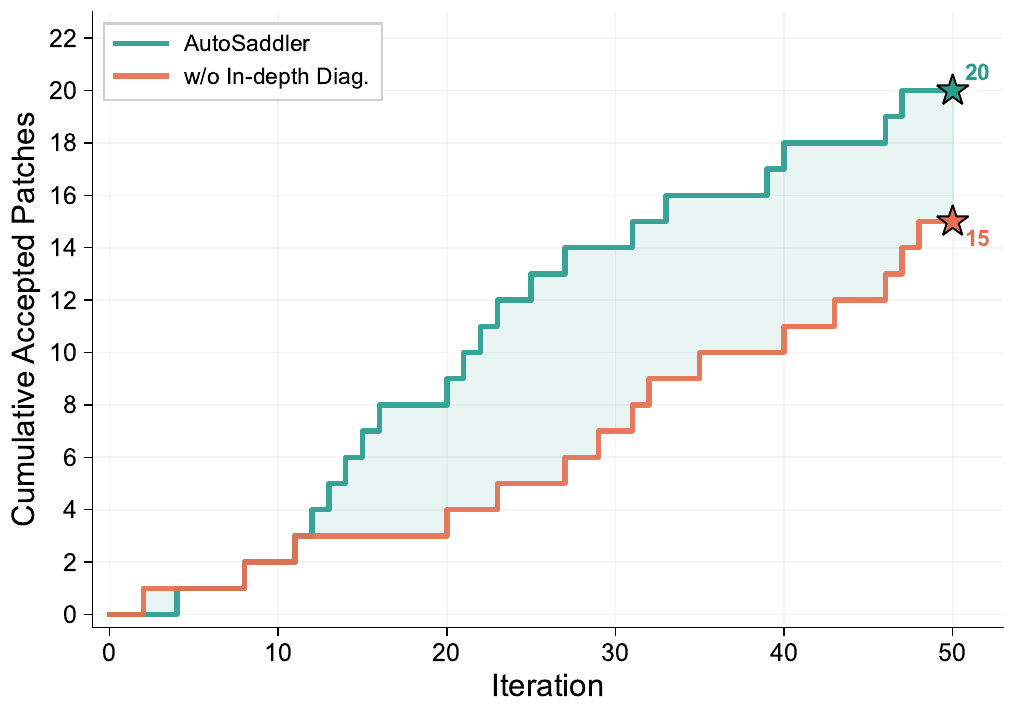}
    \caption{
        \textbf{Cumulative accepted patches.}
        \proposed{} accumulates 20 accepted patches, whereas the 
        \emph{w/o In-depth Diagnosis} ablation reaches 15.
    }
    \label{fig:rq1_accepted_patches}
\end{figure}

\paragraph{Accumulation of Accepted Patches.}
\autoref{fig:rq1_accepted_patches} visualizes the cumulative number of accepted patches over training iterations, complementing the discussion in the main text. The results show that the gap between \proposed{} and the ablation emerges early in training, by Iteration 25, and persists throughout Epoch 2. This confirms that the in-depth diagnosis component leads to more accepted patches and ultimately better performance.

\section{Additional Analysis for RQ2}
\label{app:more_details_RQ2}

\begin{figure}[!h]
\centering
    \includegraphics[width=1\linewidth]{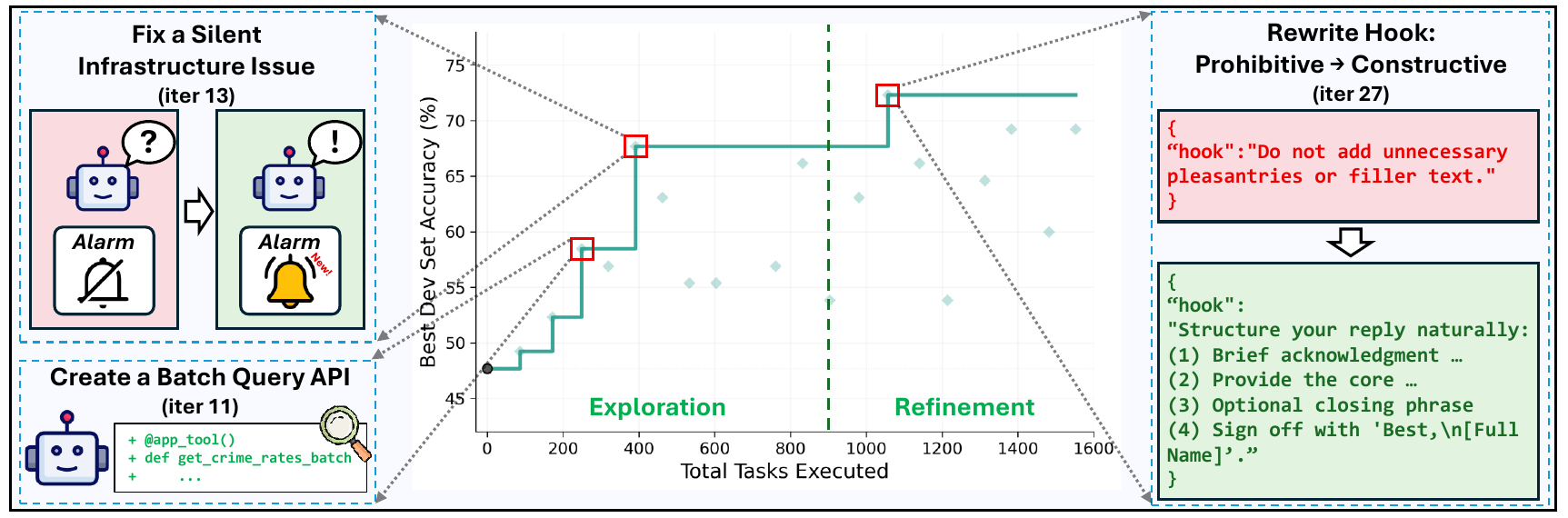}
    \caption{\textbf{Representative patches discovered over the optimization trajectory.} During the Exploration phase (Iterations $1$--$25$), the search is dominated by Capability patches, including New Tool Addition, Implementation Fix, and Agent Loop Logic Change, which address structural limitations in the harness (e.g., Iterations $11$ and $13$). During the Refinement phase (Iterations $26$--$50$), the search shifts toward Steering patches that refine prompt rules, tool descriptions, and PreToolUse hooks (e.g., Iteration $27$). Together, these two phases produce a cumulative $24.6\,pp$ improvement.}
    \label{fig:analysis_rq2}
\end{figure}

The development-set optimization trajectory in \autoref{fig:analysis_rq2} illustrates the concrete mechanism behind Structured Patching and Phased Patch Scheduling. During the Exploration phase (Iterations $1$--$25$), the search prioritizes Capability patches, which address structural limitations that cannot be resolved through prompt edits alone. For example, at Iteration $11$, \proposed{} introduces a batched API, \texttt{get\_crime\_rates\_batch}, that consolidates multiple serial queries and substantially reduces step-budget consumption. At Iteration $13$, \proposed{} identifies an execution stall in the environment's polling logic and resolves it by implementing a notification-based exit strategy.

After these capability-level improvements stabilize the harness, the Refinement phase (Iterations $26$--$50$) shifts the search toward Steering patches that fine-tune agent behavior. For instance, at Iteration $27$, \proposed{} replaces a prohibitive hook with a constructive response template, recovering cases that had previously failed due to over-correction and reaching the peak dev-set accuracy of $72.3\%$. Overall, the trajectory suggests that Structured Patching and Phased Patch Scheduling first expand the harness's functional capacity and then refine its decision-making behavior, yielding a cumulative $24.6\,pp$ improvement.

\section{Additional Analysis for RQ3}
\label{app:more_details_RQ3}

\begin{figure}[!h]
\centering
    \includegraphics[width=1\linewidth]{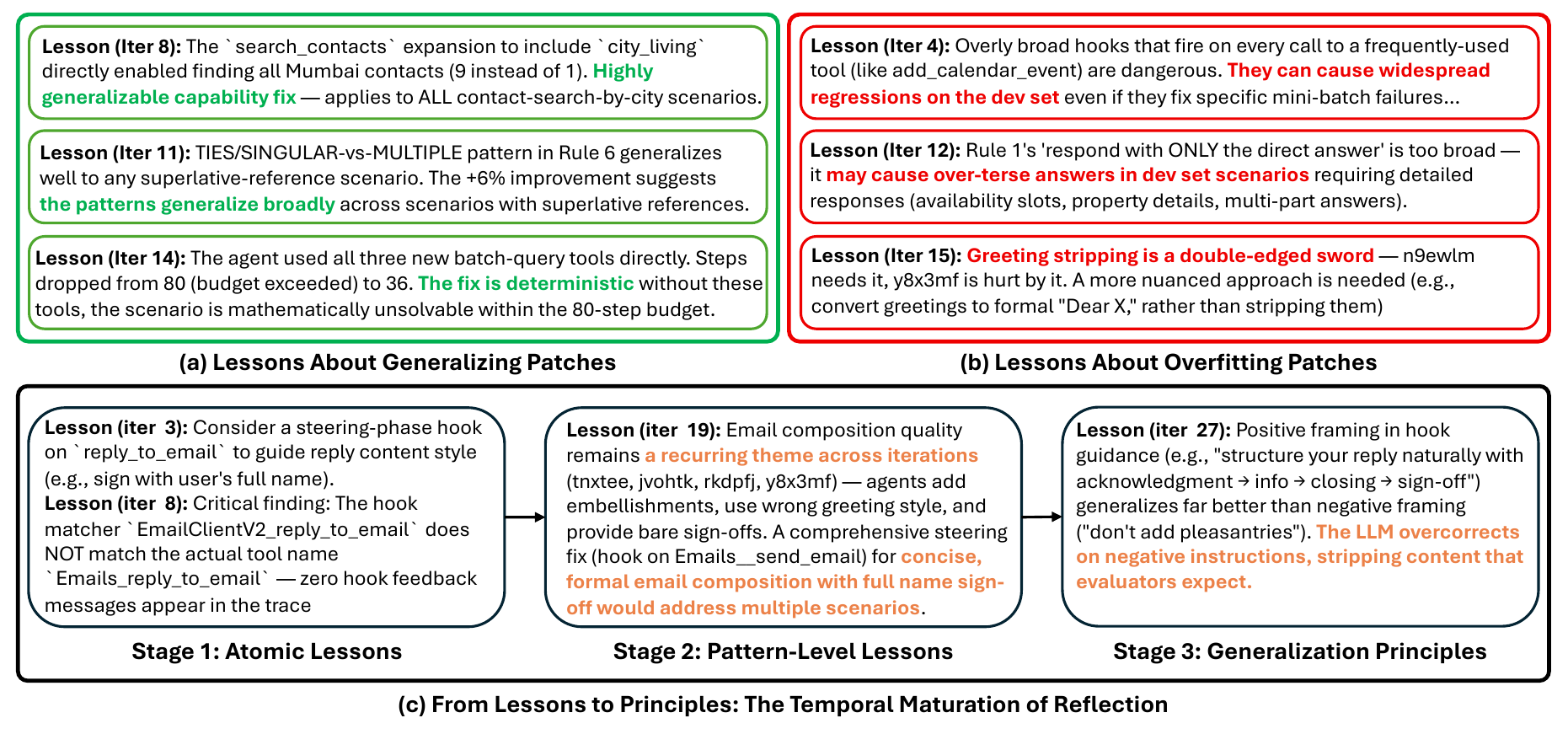}
    \caption{Reflection extracts lessons from (a)~patches that generalize and (b)~patches that overfit despite identical mini-batch outcomes. (c)~The lineage of \texttt{reply\_to\_email}-related failures illustrates how atomic lessons gradually evolve into a generalization principle.
    }
    \label{fig:rq3_reflection_lessons}
\end{figure}

We further investigate the mechanism underlying the regression gap between \proposed{} and the "\textit{w/o Generalization-Aware Selection}" ablation setting by examining the lessons extracted by reflection at each iteration (\autoref{fig:rq3_reflection_lessons}). 
Two key observations emerge. 
First, panels~(a) and (b) demonstrate that the reflection session, by jointly analyzing dev-set score fluctuations and patch contents, accurately differentiates between generalizing and overfitting patches. For example, reflection correctly endorses patches covering coherent scenario classes—such as a search-field expansion for city-based queries (Iter~8) or a tie-breaking rule for superlative references (Iter~11). Conversely, it successfully flags overfitting patches whose side effects spill over into unintended scenarios, such as a calendar hook firing indiscriminately on all event creations (Iter~4) or a conciseness rule damaging scenarios requiring structured detail (Iter~12). These accumulated judgments provide the crucial filtering signal consumed by the generalization-aware selection. Second, panel~(c) traces a single problem—email reply quality—across four Iterations and shows that reflection's knowledge of this problem changes form: from scenario-specific diagnostics (Iter~3 identifies the need for a reply hook; Iter~8 catches a matcher bug), to a pattern recognized across four scenarios (Iter~19), and finally to a scenario-independent principle on how steering text should be framed (Iter~27). What matures is not the problem itself but reflection's level of abstraction over it—evidence of cumulative learning rather than repeated rediscovery.

\section{Catalog of Discovered Patches on GAIA2}
\label{app:gaia2-patches}

\autoref{tab:gaia2-patches} presents representative patches between the
\emph{base harness} (the seed harness prior to optimization) and the
\emph{final harness} $\hat\theta_{\mathrm{AS}}$ that \proposed{} produces
after training on GAIA2. Patches are grouped by category and sub-type
following the taxonomy in \autoref{tab:patch-taxonomy}; inserted text is
highlighted in \textcolor{teal}{teal}, and italicized base-harness entries
indicate that the corresponding component or feature did not exist before
the patch.

\renewcommand{\arraystretch}{1.15}

{\fontsize{8}{9}\selectfont
\setlength{\LTpre}{\medskipamount}
\setlength{\LTpost}{\medskipamount}
\setlength{\LTcapwidth}{\linewidth}
\begin{longtable}{@{}p{0.025\linewidth} p{0.20\linewidth} p{0.345\linewidth} p{0.345\linewidth}@{}}
\caption{Representative patches discovered by \proposed{} on GAIA2,
producing the final harness $\hat\theta_{\mathrm{AS}}$ from the base harness.
C\,=\,Capability, S\,=\,Steering. Inserted text is highlighted in
\textcolor{teal}{teal}.}
\label{tab:gaia2-patches} \\*
\toprule
\textbf{C/S} & \textbf{Component} & \textbf{Base Harness} & \textbf{Final Harness ($\hat\theta_{\mathrm{AS}}$)} \\
\midrule
\endfirsthead
\multicolumn{4}{l}{\textit{(\autoref{tab:gaia2-patches} continued from previous page)}} \\
\toprule
\textbf{C/S} & \textbf{Component} & \textbf{Base Harness} & \textbf{Final Harness ($\hat\theta_{\mathrm{AS}}$)} \\
\midrule
\endhead
\midrule
\multicolumn{4}{r}{\textit{(continued on next page)}} \\
\endfoot
\bottomrule
\endlastfoot
%
%
\multicolumn{4}{@{}l}{\textbf{Prompt Patch} \quad \textit{(modifications to
\texttt{default\_agent/prompts/system\_prompt.py})}} \\
\midrule
\multicolumn{4}{@{}l}{\quad\textit{Prompt Rule Modification}} \\
\midrule
S
 & Rule 1\newline (Communication)
 & 1. COMMUNICATION: Only message the user when completely done or if the
   task is impossible.
 & 1. COMMUNICATION: Only message the user when completely done or if the
   task is impossible. \textcolor{teal}{When the user asks a factual question
   expecting a specific answer (a city name, a number, a person's name,
   etc.), respond with ONLY the direct answer~--- no extra calculations,
   context, or formatting.} \\
\addlinespace
S
 & Rule 6\newline (Ambiguity)
 & 6. AMBIGUITY: Execute all clear and unambiguous parts of a request
   immediately. When you encounter ambiguities, contradictions, or impossible
   elements, finish unambiguous subtasks and then stop and explicitly ask
   the user for clarification before proceeding with those specific parts.
 & 6. AMBIGUITY: Execute all clear and unambiguous parts \ldots\ ask the
   user for clarification before proceeding with those specific parts.
   \textcolor{teal}{A subtask is CLEAR if all its required tool parameters
   (recipients, content, IDs, paths) can be fully determined~--- even if it
   semantically relates to an ambiguous part. Common ambiguity patterns that
   REQUIRE clarification: (a)~TIES~--- when the task references `the item
   with the highest/lowest~X' but multiple items are tied at that value;
   (b)~SINGULAR vs MULTIPLE~--- when the task uses singular language (`the
   event', `the property') but multiple items match the criteria;
   (c)~PER-GROUP vs PER-ITEM~--- when applying an action `for each
   day/group' at a time derived from a singular source, but multiple sources
   exist within that group.} \\
\midrule
\multicolumn{4}{@{}l}{\quad\textit{Prompt Rule Addition}} \\
\midrule
S
 & Rule 7\newline (Task Decomposition)
 & \textit{(rule does not exist)}
 & \textcolor{teal}{7. TASK DECOMPOSITION: When the user's request contains
   multiple ordered steps (signaled by words like ``then'', ``after that'',
   ``next'', ``afterwards''), treat each step as a separate subtask.
   Complete each subtask fully before moving to the next. If a later subtask
   has ambiguities but earlier subtasks are clear, execute all clear
   subtasks first, then ask about the ambiguous ones. CRITICAL: Ordering
   words create HARD DEPENDENCY BOUNDARIES~--- subtasks after the boundary
   MUST NOT be executed until all preceding subtasks are completed, even
   if the later subtask's parameters are fully determinable.} \\
\midrule
%
%
\multicolumn{4}{@{}l}{\textbf{Tool Patch}} \\
\midrule
\multicolumn{4}{@{}l}{\quad\textit{New Tool Addition}} \\
\midrule
C
 & \texttt{CityApp.}\newline\texttt{get\_crime\_rates\_batch}
 & \textit{(tool does not exist)}
 & \texttt{get\_crime\_rates\_batch(}\newline
   \texttt{\quad zip\_codes:\,list[str])}\newline
   \texttt{\quad ->\,dict}\newline
   {\footnotesize{}Batch crime-rate lookup over many zip codes in a single
   call; replaces N serial queries that exhausted the step budget.} \\
\addlinespace
C
 & \texttt{EmailClientApp.}\newline\texttt{get\_email\_addresses\_}\newline
   \texttt{in\_folder}
 & \textit{(tool does not exist)}
 & \texttt{get\_email\_addresses\_in\_folder(}\newline
   \texttt{\quad folder\_name:\,str\,=\,"INBOX")}\newline
   \texttt{\quad ->\,dict}\newline
   {\footnotesize{}Returns all unique sender/recipient addresses in a
   folder for fast contact discovery.} \\
\addlinespace
C
 & \texttt{MessagingApp.}\newline\texttt{get\_all\_participants}
 & \textit{(tool does not exist)}
 & \texttt{get\_all\_participants()\,->\,dict}\newline
   {\footnotesize{}Returns all unique participant names across
   conversations, excluding ``Me''.} \\
\addlinespace
C
 & \texttt{MessagingAppV2.}\newline\texttt{get\_all\_participants}
 & \textit{(tool does not exist)}
 & \texttt{get\_all\_participants()\,->\,dict}\newline
   {\footnotesize{}V2 analog: returns unique participant IDs, excluding
   the current user.} \\
\addlinespace
C
 & \texttt{CabApp.}\newline\texttt{search\_ride\_history}
 & \textit{(tool does not exist; only paginated \texttt{get\_ride\_history}
   was available)}
 & \texttt{search\_ride\_history(}\newline
   \texttt{\quad date:\,str\,|\,None\,=\,None,}\newline
   \texttt{\quad start\_location\_keyword:}\newline
   \texttt{\qquad str\,|\,None\,=\,None,}\newline
   \texttt{\quad end\_location\_keyword:}\newline
   \texttt{\qquad str\,|\,None\,=\,None)\,->\,dict}\newline
   {\footnotesize{}Searches the full ride history with optional filters in
   one call.} \\
\midrule
\multicolumn{4}{@{}l}{\quad\textit{Implementation Fix}} \\
\midrule
C
 & \texttt{ContactsApp.}\newline\texttt{search\_contacts}
 & Searches \texttt{first\_name},\newline
   \texttt{last\_name},
   \texttt{phone\_number},\newline\texttt{email} only.
 & Additionally searches\newline
   \textcolor{teal}{\texttt{city\_living}},
   \textcolor{teal}{\texttt{country}},\newline
   \textcolor{teal}{\texttt{nationality}},
   \textcolor{teal}{\texttt{job}},
   \textcolor{teal}{\texttt{address}},\newline
   enabling city- and role-based contact lookup. \\
\midrule
%
%
\multicolumn{4}{@{}l}{\textbf{Middleware Patch}} \\
\midrule
\multicolumn{4}{@{}l}{\quad\textit{PreToolUse Hook}} \\
\midrule
S
 & \texttt{Emails\_\_}\newline\texttt{reply\_to\_email}\newline
   {\footnotesize{}(reminder text in \texttt{hook.json})}
 & ``Do not add unnecessary pleasantries or filler text.''\newline
   {\footnotesize{}(prohibitive: triggers over-correction~--- agent strips
   content evaluators expect.)}
 & \textcolor{teal}{``Structure your reply naturally: (1)~brief
   acknowledgment, (2)~provide the core information, (3)~optional closing
   phrase, (4)~sign off with `Best,\textbackslash n[Full Name]'.''}\newline
   {\footnotesize{}(constructive: provides a positive structural template.)} \\
\midrule
\multicolumn{4}{@{}l}{\quad\textit{Agent Loop Logic Change}} \\
\midrule
C
 & \texttt{are\_simulation\_}\newline\texttt{main.py}\newline
   {\footnotesize{}(agent loop)}
 & The benchmark environment does not notify the agent of cab status
   changes (e.g., delays). Without a mechanism to advance simulated time
   between turns, the agent loop stalls indefinitely, unable to reach the
   point where it could proactively discover these state changes.
 & \proposed{} resolves the stall by inserting an
   \textcolor{teal}{inter-turn time-advancement call}: when no messages are
   pending, the loop invokes
   \textcolor{teal}{\texttt{wait\_for\_notification(}}\newline
   \textcolor{teal}{\texttt{timeout=300)}} to fast-forward simulated time by
   up to 300 seconds. If no notification arrives, the timeout itself wakes
   the agent, prompting it to proactively poll for state changes. \\
\addlinespace
C
 & \texttt{\_resolve\_fs\_path}\newline
   in \texttt{execute\_}\newline\texttt{tool\_call()}\newline
   on \texttt{*\_\_mkdir} (\texttt{path}),\newline
   \texttt{*\_\_mv} (\texttt{path2})
 & \texttt{path="benchmarks/sims3"}\newline
   {\footnotesize{}(no automatic resolution: \texttt{mkdir} silently creates
   a new \texttt{benchmarks/sims3/} at the filesystem root, despite an
   existing \texttt{Documents/benchmarks/} elsewhere.)}
 & \texttt{path=\textcolor{teal}{"Documents/}}\newline
   \texttt{\textcolor{teal}{benchmarks/sims3"}}\newline
   {\footnotesize{}\textcolor{teal}{(automatic path resolution: locates
   existing parent directories with the same basename via filesystem search
   before \texttt{mkdir}/\texttt{mv} is invoked.)}} \\
\end{longtable}
}


\section{Catalog of Discovered Patches on Terminal-Bench 2.0}
\label{app:tb2-patches}

\autoref{tab:tb2-patches} presents representative
patches between the \emph{base harness} (the seed harness prior to
optimization) and the \emph{final harness}
$\hat\theta_{\mathrm{AS}}$ that \proposed{} produces after training on
Terminal-Bench~2.0. Patches are grouped by category and sub-type
following the taxonomy in~\autoref{tab:patch-taxonomy}; inserted text is
highlighted in \textcolor{teal}{teal}, and italicized base-harness entries
indicate that the corresponding component or feature did not exist before
the patch.

\renewcommand{\arraystretch}{1.15}

{\fontsize{8}{9}\selectfont
\setlength{\LTpre}{\medskipamount}
\setlength{\LTpost}{\medskipamount}
\setlength{\LTcapwidth}{\linewidth}
\begin{longtable}{@{}p{0.025\linewidth} p{0.20\linewidth} p{0.345\linewidth} p{0.345\linewidth}@{}}
\caption{Representative patches discovered by \proposed{}
on Terminal-Bench~2.0, producing the final harness
$\hat\theta_{\mathrm{AS}}$ from the base harness.
C\,=\,Capability, S\,=\,Steering. Inserted text is highlighted in
\textcolor{teal}{teal}.}
\label{tab:tb2-patches} \\*
\toprule
\textbf{C/S} & \textbf{Component} & \textbf{Base Harness} & \textbf{Final Harness ($\hat\theta_{\mathrm{AS}}$)} \\
\midrule
\endfirsthead
\multicolumn{4}{l}{\textit{(\autoref{tab:tb2-patches} continued from previous page)}} \\
\toprule
\textbf{C/S} & \textbf{Component} & \textbf{Base Harness} & \textbf{Final Harness ($\hat\theta_{\mathrm{AS}}$)} \\
\midrule
\endhead
\midrule
\multicolumn{4}{r}{\textit{(continued on next page)}} \\
\endfoot
\bottomrule
\endlastfoot
%
%
\multicolumn{4}{@{}l}{\textbf{Prompt Patch} \quad \textit{(modifications to
the system prompt template
\texttt{prompt-templates/verify-before-complete.txt})}} \\
\midrule
\multicolumn{4}{@{}l}{\quad\textit{Prompt Rule Addition}} \\
\midrule
S
 & Verification\newline before completion
 & \textit{(no verification guidance;\newline the base prompt ends
   immediately after the format specification)}
 & \textcolor{teal}{CRITICAL -- Verification before completion:
   You MUST verify your solution works before setting
   \texttt{"task\_complete":\,true}. When setting up services, test
   endpoints. When creating files, verify contents. Verify file permissions
   for all service users. Perform a dry run of automated hooks. If
   verification fails, fix and verify again before completing.} \\
\addlinespace
S
 & Different-input\newline testing \&\newline build-artifact\newline
   clean-up
 & \textit{(rule does not exist)}
 & \textcolor{teal}{For scripts or programs, test with a DIFFERENT input
   than the provided example to confirm generalizability. After testing,
   clean up the output directory: remove any build artifacts (compiled
   binaries, \texttt{.o} files), temporary files, or intermediate outputs
   that the task did NOT ask you to create. Verifiers often check that
   output directories contain ONLY the expected files.} \\
\addlinespace
S
 & Library\newline preference
 & \textit{(rule does not exist)}
 & \textcolor{teal}{When writing scripts, check the terminal output for
   pre-installed packages and prefer established libraries over fragile
   custom implementations.} \\
\addlinespace
S
 & HTML\newline sanitization
 & \textit{(rule does not exist)}
 & \textcolor{teal}{For HTML sanitization / XSS filtering, ALWAYS use
   BeautifulSoup (not regex). Decompose dangerous tags (\texttt{script},
   \texttt{noscript}, \texttt{iframe}, \texttt{object}, \texttt{embed},
   \texttt{frame}, \texttt{style}) and do a string replacement to strip
   \texttt{javascript:} from entire output as final safety pass.} \\

\midrule
%
%
\multicolumn{4}{@{}l}{\textbf{Tool Patch}} \\
\midrule
\multicolumn{4}{@{}l}{\quad\textit{Argument Modification}} \\
\midrule
C
 & \texttt{\_limit\_output\_}\newline\texttt{length}\newline
   {\footnotesize{}(\texttt{max\_bytes}\newline parameter)}
 & \texttt{\_limit\_output\_length(}\newline
   \texttt{\quad output,\;max\_bytes=10000)}\newline
   {\footnotesize{}Default 10\,KB limit truncates large file contents or
   command outputs, causing the agent to miss critical information.}
 & \texttt{\_limit\_output\_length(}\newline
   \texttt{\quad output,\;max\_bytes=\textcolor{teal}{30000})}\newline
   {\footnotesize{}\textcolor{teal}{30\,KB limit} provides more room while
   keeping context manageable.} \\
\midrule
\multicolumn{4}{@{}l}{\quad\textit{Implementation Fix}} \\
\midrule
C
 & \texttt{\_get\_prompt\_}\newline\texttt{template\_path}\newline
   {\footnotesize{}(prompt template\newline selection)}
 & Returns the \textbf{default}\newline Terminus-2 prompt template path.\newline
   {\footnotesize{}The default prompt contains only JSON format
   specification without domain-specific guidance.}
 & Returns path to\newline
   \textcolor{teal}{\texttt{verify-before-complete.txt}}\newline
   {\footnotesize{}\textcolor{teal}{Custom prompt template} that includes
   verification guidance, library-preference rules, and coding best
   practices alongside the standard format specification.} \\
\midrule
%
%
\multicolumn{4}{@{}l}{\textbf{Middleware Patch}} \\
\midrule
\multicolumn{4}{@{}l}{\quad\textit{PreToolUse Hook}} \\
\midrule
S
 & \texttt{\_get\_completion\_}\newline\texttt{confirmation\_}\newline
   \texttt{message}\newline
   {\footnotesize{}(completion\newline confirmation)}
 & ``Current terminal state:\textbackslash n\{output\}\textbackslash n
   \textbackslash n Are you sure you want to mark the task as complete?
   This will trigger your solution to be graded and you won't be able to
   make any further corrections.\,''\newline
   {\footnotesize{}(generic confirmation; agent often skips verification)}
 & ``Current terminal state:\textbackslash n\{output\}\textbackslash n
   \textbackslash n \textcolor{teal}{STOP -- You must verify your solution
   before confirming completion.\textbackslash n For service tasks: test
   endpoints, verify permissions.\textbackslash n For coding tasks: test
   with a DIFFERENT input, verify output structure.\textbackslash n Clean
   up: remove build artifacts the task did NOT request.}''\newline
   {\footnotesize{}(\textcolor{teal}{structured verification checklist}
   with task-type-specific guidance)} \\
\midrule
\multicolumn{4}{@{}l}{\quad\textit{Agent Loop Logic Change}} \\
\midrule
C
 & \texttt{run()}\newline
   {\footnotesize{}(environment\newline discovery)}
 & Agent starts with no knowledge of what packages are pre-installed in the
   Docker container. Frequently chooses fragile custom implementations
   (e.g., regex for HTML) over established libraries that are already
   available.
 & Before the agent loop begins, \textcolor{teal}{\texttt{pip\;list
   {-}{-}format=columns}} is executed in the container so the initial
   terminal state shows all pre-installed Python packages. The agent can
   then choose appropriate libraries (e.g., BeautifulSoup over regex)
   from the start. \\
\end{longtable}
}

\section{Instructions Used In \proposed{}}
\label{sec:instructions_used_in_autosaddler}

To ensure reproducible and principled optimization, \proposed{} employs a multi-session orchestration design where each session is guided by a specialized prompt template. These instructions are designed to operationalize the core pillars of our framework: in-depth debugging, structured harness intervention, and generalization-aware selection.

The \textbf{Diagnosis-Patch Session} (\autoref{fig:Diagnosis-Patch-session-prompt}) initiates the optimization loop by grounding the agent in the current mini-batch failures. Unlike simple reflection-based approaches, this prompt enforces a rigorous "diagnose-before-patch" workflow. It provides the agent with access to raw execution traces alongside the harness codebase and mandates the use of specialized skills (e.g., \texttt{history-analysis}, \texttt{diagnose}) to identify the underlying root causes. Furthermore, it implements our phased scheduling by restricting the agent to either \textit{capability} or \textit{steering} patches, ensuring that interventions are targeted and architecturally sound.

The \textbf{Reflection Session} (\autoref{fig:reflection-session-prompt}) focuses on causal attribution and the extraction of durable lessons. The prompt instructs the agent to perform a comparative analysis of pre- and post-patch execution traces for every scenario in the mini-batch. A key feature of this session is the mandatory "stochasticity check", where the agent must distinguish between true causal fixes/regressions and stochastic artifacts caused by LLM non-determinism. By categorizing outcomes into four states (\textit{fixed, regressed, still-failing, still-passing}) and recording detailed evidence in the EvoDAG, this session ensures that only verified insights drive the evolution process.

Finally, the \textbf{Evolution Session} (\autoref{fig:evolution-session-prompt}) serves as the framework's meta-optimizer. It provides the agent with a global view of the EvoDAG history, including accumulated lessons, performance metrics across development sets, and code diffs from prior iterations. The prompt guides the agent to synthesize the next harness candidate not just by incremental editing, but by potentially merging successful components from different lineages or reverting regressive updates. This session ensures that the resulting harness base is stable, verified for syntax and logic, and optimized for broad generalization across the task distribution.

\begin{tcolorbox}[
    enhanced,
    breakable,                       
    colback=white,
    colframe=gray!50,
    boxrule=0.4pt,
    arc=4pt,
    left=8pt, right=8pt, top=4pt, bottom=4pt,
    title=\textbf{Diagnosis-Patch Prompt},
    coltitle=white,
    colbacktitle=black,
    fonttitle=\small\bfseries,
    halign title=center,
    attach boxed title to top={yshift=-0.5mm, xshift=-0.4pt},
    boxed title style={
        colback=black,
        colframe=black,
        boxrule=0pt,
        arc=4pt,
        sharp corners=south,
    },
    width=\linewidth,
    width=\linewidth,
    pad at break*=2mm,
    overlay middle and last={
        \node[anchor=south east, font=\footnotesize\itshape, gray]
        at ([xshift=-6pt,yshift=2pt]frame.north east)
        {(continued)};
    },
]
{\fontsize{8pt}{9.5pt}\ttfamily%
\setlength{\parindent}{0pt}%
\setlength{\parskip}{2pt}%
\raggedright%
\obeylines%
\#\# Mandatory Skills
 
You MUST read and follow the SKILL.md for each skill listed below. Do NOT skip or summarize any skill --- execute the full procedure described in each one. These skills are installed at `.claude/skills/<n>/SKILL.md` in the current worktree.
 
| Order | Skill | When | Why |
|-------|-------|------|-----|
| 1 | `history-analysis` | \textbf{Before any other work} (Step 1) | Structured analysis of the full evolution history --- identifies relevant lessons, prior attempts, proven strategies, and regression-prone areas |
| 2 | `diagnose` | \textbf{Before applying any patch} (Step 2) | Root-cause analysis from agent traces and codebase --- pinpoints exactly why each failing scenario fails |
| 3 | `capability-patch` or `steering-patch` | \textbf{When applying patches} (Step 3) | Phase-appropriate patch methodology --- `capability-patch` during capability phase, `steering-patch` during steering phase. Follow the skill matching the current phase |
| 4 | `patch-verification` | \textbf{After applying all patches} (Step 4) | Crash-safety verification --- syntax, imports, docstrings, hooks, logic. A crashing patch is worse than no patch |
 
> \textbf{Enforcement}: Every step above is mandatory and sequential. Do NOT apply patches without first completing diagnosis. Do NOT finish the session without running the full `patch-verification` procedure. Do NOT skip `history-analysis` --- it prevents repeating known failures.
 
\#\# Goal
 
Fix failing scenarios in this mini-batch by diagnosing root causes from execution traces and the agent codebase, then applying targeted code patches --- without regressing scenarios that already pass.
 
Each scenario is a task the agent must solve. The initial evaluation on the mini-batch has already run the agent on each scenario and recorded its execution trace: every tool call, every response, and every reasoning step. Your job is to read these traces alongside the agent codebase, identify \textbf{why} the agent failed, determine what code change would fix the root cause, and apply it.
 
The optimization objective is to \textbf{maximize performance across the full task distribution} --- not just this mini-batch. Patches should address general behavioral patterns, not hardcode scenario-specific answers. A patch that fixes one scenario by adding a general rule will also help similar unseen scenarios; a patch that hardcodes a specific answer helps only that one scenario and may harm others.
 
\textbf{Acceptance}: A patch is accepted if the sum of re-evaluation scores exceeds the sum of initial scores. Even fixing a single scenario is enough. However, regressions (passing $\rightarrow$ failing) are recorded as bad patterns --- avoid them.
 
\#\# Context
 
- \textbf{Iteration}: \{iteration\}
- \textbf{Candidate}: C\{candidate\_idx\}
- \textbf{Current worktree}: `\{worktree\_path\}`
- \textbf{Base parent}: C\{base\_parent\_idx\} (worktree: `\{parent\_worktree\}`)
\{cherry\_pick\_parents\_section\}- \textbf{Phase}: \{phase\}
- \textbf{Initial evaluation output}: `\{before\_output\_dir\}`
 
\#\# Current Mini-Batch (\{num\_scenarios\} scenarios)
 
\{mini\_batch\_listing\}
 
\#\#\# Initial Scores (pass rate: \{before\_pass\_rate\})
 
\{before\_scores\_listing\}
 
\#\# Patch Types for This Phase
 
\{patch\_types\_section\}
 
\#\# Workflow
 
\#\#\# 1. Analyze history
 
\textbf{MANDATORY first step.} Run the `history-analysis` skill to build a structured understanding of the full evolution history. Do NOT skip this step or truncate `evo-dag show history` with `head`/`tail`. The skill provides the methodology for reading the complete history without truncation. Focus on:
- Relevant lessons (good/bad patterns) for the current mini-batch
- Prior attempts on the failing scenarios --- what worked and what failed
- Proven patch strategies and regression-prone areas
 
\#\#\# 2. Diagnose failing scenarios
 
For each failing scenario, use the `diagnose` skill to identify the root cause. The key is to find \textbf{why} the agent made the wrong decision, not just \textbf{what} went wrong. Diagnosis requires reading both the execution trace and the agent codebase.
 
\textbf{Trace files} (output dir: `\{before\_output\_dir\}`): See CLAUDE.md's \textbf{Iteration Output Structure} section for the exact file layout. Key files to read:
- \textbf{Evaluation rationale}: Per-scenario scores and judge rationale --- the evaluator's summary of what was expected vs what the agent produced.
- \textbf{Agent execution traces}: Per-scenario full tool call, response, and reasoning traces --- where you find the exact point where the agent diverged from correct behavior.
 
\textbf{Agent codebase}: current worktree `\{worktree\_path\}`
 
The `diagnose` skill provides the full methodology --- follow it for each failing scenario.
 
\#\#\# 3. Apply patches
 
Based on your diagnosis, use the phase-appropriate patch skill:
- \textbf{Capability phase} (`capability-patch`): Expand what the agent CAN DO --- new tool methods, parameter additions, implementation fixes, infrastructure changes. These unlock scenarios that are unreachable through prompt tuning alone.
- \textbf{Steering phase} (`steering-patch`): Refine HOW the agent behaves --- prompt rules, tool description corrections, PreToolUse hooks. These fine-tune the agent's use of existing capabilities.
 
\textbf{Patch guidelines}:
- \textbf{Target the root cause, not the symptom.} If the agent sends a wrong email subject, the fix is a general rule about deriving subjects from user wording --- not hardcoding the correct subject for one scenario.
- \textbf{Keep patches generalizable.} Rules should be abstract: no scenario IDs, no specific names, no hardcoded answers. A good patch helps all scenarios that share the same root cause pattern.
- \textbf{Align agent-facing text with capability changes.} When you add or modify tools, parameters, or infrastructure code, also update the system prompt, tool docstrings, and/or hooks so the agent is aware of the changes. Check for conflicting rules.
- \textbf{Learn from patch history.} Check `evo-dag show history` --- it contains diffs, per-scenario results, reflections, and accumulated lessons. Use it to understand: which patches generalized well to the dev set, which caused regressions, which root-cause patterns were effectively resolved, and which approaches repeatedly failed. Build on proven strategies and avoid repeating known bad patterns.
- \textbf{Prefer replacement over addition} when modifying prompts. The system prompt has a finite attention budget --- adding rules without removing or merging existing ones dilutes their impact.
 
\#\#\# 4. Verify patches
 
After applying all patches, you MUST run the `patch-verification` skill to ensure the patched codebase has no runtime errors. A patch that crashes at runtime is worse than no patch --- verification is mandatory before committing. Do NOT skip this step under any circumstances.
 
\#\#\# 5. Write reasoning and record intent
 
After applying all patches, write your reasoning to `proposer\_reasoning.md` in the working directory root. For each targeted scenario, include: task description, expected vs actual behavior, trace analysis (which step went wrong), root cause, and resolution strategy (generalizable rule). This sharpens the `--diagnosis` that propagates to future iterations via `evo-dag show history`. Then record intent:
\ \ \ \ ```bash
\ \ \ \ evo-dag update-intent \textbackslash
\ \ \ \ \ \ --target-scenarios "id1,id2" \textbackslash
\ \ \ \ \ \ --diagnosis "Root cause analysis of why target scenarios fail" \textbackslash
\ \ \ \ \ \ --approach "Brief description of the patch approach" \textbackslash
\ \ \ \ \ \ --files-changed "f1.py,f2.py" \textbackslash
\ \ \ \ \ \ --change-summary "What was changed and why"
\ \ \ \ ```
 
\textbf{CRITICAL}: You MUST run `evo-dag update-intent` before finishing this session. Failure to do so will result in incomplete iteration records and degrade the quality of future iterations' patch history analysis.
 
\#\# What Happens Next
 
After this session, the outer loop re-evaluates the patched worktree on the same mini-batch, computes initial/re-evaluation per-scenario impacts (fixed, regressed, still\_failing, still\_passing), and records the patch verdict. If the patch is accepted (re-evaluation score > initial score), the candidate is also evaluated on the full dev-set to measure generalizability. Then \textbf{Session 2 (Reflection)} runs, where you analyze the results and record structured reflections for each scenario.
}
\end{tcolorbox}
 
\captionof{figure}{Diagnosis-Patch Session Prompt: agent receives the current mini-batch's failing scenarios with execution traces, diagnoses root causes via the agent codebase, and applies targeted, generalizable patches to the existing harness.}
\label{fig:Diagnosis-Patch-session-prompt}

\begin{tcolorbox}[
    enhanced,
    breakable,                       
    colback=white,
    colframe=gray!50,
    boxrule=0.4pt,
    arc=4pt,
    left=8pt, right=8pt, top=4pt, bottom=4pt,
    title=\textbf{Reflection Session Prompt},
    coltitle=white,
    colbacktitle=black,
    fonttitle=\small\bfseries,
    halign title=center,
    attach boxed title to top={yshift=-0.5mm, xshift=-0.4pt},
    boxed title style={
        colback=black,
        colframe=black,
        boxrule=0pt,
        arc=4pt,
        sharp corners=south,
    },
    width=\linewidth,
    width=\linewidth,
    pad at break*=2mm,
    overlay middle and last={
        \node[anchor=south east, font=\footnotesize\itshape, gray]
        at ([xshift=-6pt,yshift=2pt]frame.north east)
        {(continued)};
    },
]
{\fontsize{8pt}{9.5pt}\ttfamily%
\setlength{\parindent}{0pt}%
\setlength{\parskip}{2pt}%
\raggedright%
\obeylines%
\#\# Mandatory Skills
 
You MUST read and follow the SKILL.md for each skill listed below. Do NOT skip or summarize any skill --- execute the full procedure described in each one. These skills are installed at `.claude/skills/<n>/SKILL.md` in the current worktree.
 
| Order | Skill | When | Why |
|-------|-------|------|-----|
| 1 | `history-analysis` | \textbf{Before any other work} (Step 1) | Structured analysis of the full evolution history --- provides historical context for each scenario, pattern evolution, and dev score attribution needed for meaningful reflections |
| 2 | `diagnose` | \textbf{For every regressed AND fixed scenario} (Step 3) | Causal attribution --- determines whether each state change (PASS$\leftrightarrow$FAIL) was truly caused by the patch or is a stochastic artifact from LLM non-determinism. Accurate classification is critical to avoid polluting lessons with false signal |
 
> \textbf{Enforcement}: Do NOT attribute any state change to "stochastic noise" or "LLM non-determinism" without running the full `diagnose` skill procedure. Every fixed/regressed classification must cite specific evidence from the diagnosis.
 
\#\# Goal
 
Analyze the initial/re-evaluation results for every scenario in the mini-batch and record structured reflections that future iterations can learn from.
 
Your reflections feed directly into `evo-dag show history` --- the primary knowledge store that future iterations consult when diagnosing failures, choosing patch strategies, and avoiding past mistakes. The quality of your reflections determines how effectively the pipeline learns. Be specific: reference exact tool calls, code paths, and behavioral patterns. Vague reflections waste learning potential.
 
\#\# Context
 
- \textbf{Iteration}: \{iteration\}
- \textbf{Candidate}: C\{candidate\_idx\}
- \textbf{Current worktree}: `\{worktree\_path\}`
- \textbf{Base parent}: C\{base\_parent\_idx\} (worktree: `\{parent\_worktree\}`)
- \textbf{Phase}: \{phase\}
- \textbf{Initial evaluation output}: `\{before\_output\_dir\}`
- \textbf{Re-evaluation output}: `\{train\_after\_cycle\_dir\}`
 
\#\# Session 1 Reasoning
 
The following is the proposer's diagnosis and patch rationale from Session 1. Use this to understand the \textbf{intent} behind the patch --- what root causes were identified, what the patch was designed to fix, and what behavioral changes were expected.
 
\{proposer\_reasoning\}
 
\#\# Results Summary
 
\{results\_summary\}
 
\#\# Per-Scenario Details
 
\{per\_scenario\_details\}
 
\{generalization\_section\}
 
\#\# The Four Outcomes
 
| Before | After | Status | Key Question |
|--------|-------|--------|-------------|
| FAIL | PASS | `fixed` | What root cause did the patch resolve? How? |
| PASS | FAIL | `regressed` | What did the patch break? Why? |
| FAIL | FAIL | `still\_failing` | Why was the patch insufficient? What to try next? |
| PASS | PASS | `still\_passing` | Did the patch interact with this scenario at all? |
 
\#\# Workflow
 
\#\#\# 1. Analyze history
 
\textbf{MANDATORY first step.} Run the `history-analysis` skill to build a structured understanding of the full evolution history. Do NOT skip this step or truncate `evo-dag show history` with `head`/`tail`. The skill provides the methodology for reading the complete history without truncation. Focus on:
- Historical context for each scenario in the mini-batch
- Pattern evolution --- emerging patterns not yet captured in lessons
- Dev score attribution --- which patches improved/hurt dev accuracy and why
 
This context is essential for writing meaningful reflections --- you need to know what was tried before to explain why this iteration's results differ, and to write `--prevention-or-next` guidance that adds new information rather than repeating existing lessons.
 
\#\#\# 2. Read traces (by priority)
 
Before traces are in `\{before\_output\_dir\}`. After traces are in `\{train\_after\_cycle\_dir\}`. See CLAUDE.md's \textbf{Iteration Output Structure} section for the exact file layout within each output directory.
 
Compare before and after traces to understand what the patch changed in the agent's behavior.
 
\textbf{Priority order}:
1. \textbf{Regressed} --- always read both before and after traces. You must understand what behavior was correct before and what the patch broke.
2. \textbf{Still-failing (targeted)} --- read to check for partial progress. Did the failure point shift? Is the agent closer to correct behavior?
3. \textbf{Fixed} --- skim the after trace to confirm the patch resolved the root cause as intended, not through a lucky side effect.
4. \textbf{Still-passing} --- skip unless the scenario uses the same tools or components you modified.
 
\#\#\# 3. Diagnose state-changed scenarios (MANDATORY)
 
\textbf{For every regressed AND fixed scenario}, run the full `diagnose` skill to determine whether the state change was \textbf{truly caused by the patch} or is a \textbf{stochastic artifact} (caused by LLM non-determinism).
 
LLM non-determinism means the same agent code can produce different tool-call sequences across runs. This affects both directions:
- A \textbf{regressed} scenario (PASS$\rightarrow$FAIL) may have nothing to do with the patch --- the agent simply took a different reasoning path.
- A \textbf{fixed} scenario (FAIL$\rightarrow$PASS) may not be a real fix --- the agent may have succeeded by luck, not because of the patch. Recording this as a true fix pollutes `good\_patterns` with false signal.
 
Accurate causal attribution is critical: false regressions accumulate noise in `bad\_patterns`, and false fixes accumulate noise in `good\_patterns`. Both degrade the quality of lessons for future iterations.
 
\textbf{How to distinguish true vs. stochastic state changes:}
 
1. \textbf{Read the before trace}: Identify the exact tool-call sequence and reasoning steps that led to the before result.
2. \textbf{Read the after trace}: Identify where the agent's behavior diverged.
3. \textbf{Check the code diff} (`evo-dag show edge \{base\_parent\_idx\} \{candidate\_idx\}`): Does the diff touch any code path, prompt text, hook, or tool that the scenario exercises? If the diff is completely unrelated to the scenario's tool usage and divergence point, it is likely stochastic.
4. \textbf{Check the divergence point}: Is the behavioral change at a step that the patch modified (true causation) or at an unrelated step where the agent simply made a different LLM-driven choice (stochastic)?
 
\textbf{Classification criteria:}
- \textbf{True (regression or fix)}: The diff modifies code/prompt/hook that the scenario directly exercises, AND the after trace shows the agent behaving differently at the modified point in a way that explains the outcome change.
- \textbf{Stochastic}: The diff does NOT touch anything the scenario exercises, OR the agent's divergence point is unrelated to the patch (e.g., different search query phrasing, different email wording).
- \textbf{Uncertain}: The diff touches a shared component but the causal link is unclear. Record as uncertain with specific evidence.
 
\textbf{Do NOT} attribute any state change to "LLM non-determinism" or "stochastic noise" without performing the above analysis. Every classification must cite:
- The specific divergence point in the traces
- Whether the diff touches the relevant code path
- The evidence for or against a causal link
 
Record the classification in `--explanation` and `--prevention-or-next`. For true regressions, explain what the patch broke. For stochastic regressions, note the evidence so future iterations don't over-correct. For true fixes, explain the causal chain from patch to fix. For stochastic fixes, note the evidence so future iterations don't over-rely on the patch strategy.
 
\#\#\# 4. Inspect code changes
 
Run `evo-dag show edge \{base\_parent\_idx\} \{candidate\_idx\}` to see the code diff, files changed, and per-scenario impacts. Cross-reference the diff with the trace behavior to understand causality.
 
For deeper inspection, read source files directly in the current worktree `\{worktree\_path\}`.
 
\#\#\# 5. Record reflections
 
Use `evo-dag update-reflection` for \textbf{every} scenario in the mini-batch.
 
\textbf{Fields}:
- `--root-cause`: The underlying reason the scenario fails (independent of the patch). What is the agent doing wrong and why?
- `--explanation`: For `fixed` --- how the patch resolved it. For `still\_failing` --- why the patch did NOT work (what was insufficient). For `regressed` --- what the patch broke and how. For `still\_passing` --- why the patch did not interfere (only for scenarios using modified components).
- `--prevention-or-next`: What to try next, or what to avoid. This is critical for all non-passing outcomes --- it creates the lessons that future iterations rely on.
- `--generalization-note`: How this scenario's outcome relates to development set accuracy. Did the patch generalize beyond the mini-batch? Record when dev scores are available.
 
\textbf{Per outcome}:
 
\textbf{`fixed`} --- You MUST have completed the diagnosis step (step 3) before recording this reflection. Classify as true fix or stochastic artifact.
\ \ \ \ ```bash
\ \ \ \ \# True fix (patch caused the success):
\ \ \ \ evo-dag update-reflection \textbackslash
\ \ \ \ \ \ --node \{candidate\_idx\} \textbackslash
\ \ \ \ \ \ --scenario "<id>" --status "fixed" \textbackslash
\ \ \ \ \ \ --root-cause "Agent called X with wrong param Y because docstring said Z." \textbackslash
\ \ \ \ \ \ --explanation "TRUE FIX: Patch corrected docstring to clarify Y. Agent now calls correctly at step 4 --- directly caused by diff in tools/email.py L42." \textbackslash
\ \ \ \ \ \ --prevention-or-next "For similar tool-misuse scenarios, check docstring accuracy first."
 
\ \ \ \ \# Stochastic artifact (not caused by the patch):
\ \ \ \ evo-dag update-reflection \textbackslash
\ \ \ \ \ \ --node \{candidate\_idx\} \textbackslash
\ \ \ \ \ \ --scenario "<id>" --status "fixed" \textbackslash
\ \ \ \ \ \ --root-cause "Agent happened to choose correct search query in after trace." \textbackslash
\ \ \ \ \ \ --explanation "STOCHASTIC: Diff only touches calendar tool. This scenario uses search tool exclusively. Agent succeeded because it picked a better search query by chance at step 2 --- unrelated to patch." \textbackslash
\ \ \ \ \ \ --prevention-or-next "Not a reliable fix --- scenario may fail again. Root cause (weak search strategy) still needs addressing."
\ \ \ \ ```
 
\textbf{`regressed`} (highest priority) --- You MUST have completed the diagnosis step (step 3) before recording this reflection. Classify as true regression or stochastic artifact with evidence.
\ \ \ \ ```bash
\ \ \ \ \# True regression (patch caused the failure):
\ \ \ \ evo-dag update-reflection \textbackslash
\ \ \ \ \ \ --node \{candidate\_idx\} \textbackslash
\ \ \ \ \ \ --scenario "<id>" --status "regressed" \textbackslash
\ \ \ \ \ \ --root-cause "This scenario relied on optional param Y in tool X." \textbackslash
\ \ \ \ \ \ --explanation "TRUE REGRESSION: Docstring change removed note about Y being optional. Agent stopped passing Y. Divergence at step 5 where agent no longer passes Y --- directly caused by diff in tools/email.py L42." \textbackslash
\ \ \ \ \ \ --prevention-or-next "When modifying docstrings, preserve param optionality annotations."
 
\ \ \ \ \# Stochastic artifact (not caused by the patch):
\ \ \ \ evo-dag update-reflection \textbackslash
\ \ \ \ \ \ --node \{candidate\_idx\} \textbackslash
\ \ \ \ \ \ --scenario "<id>" --status "regressed" \textbackslash
\ \ \ \ \ \ --root-cause "Agent used different search query phrasing in after trace." \textbackslash
\ \ \ \ \ \ --explanation "STOCHASTIC: Diff only touches calendar tool docstring. This scenario uses email tool exclusively. Agent diverged at step 3 with different query wording --- unrelated to patch." \textbackslash
\ \ \ \ \ \ --prevention-or-next "No action needed --- stochastic noise. Do not over-correct for this scenario."
\ \ \ \ ```
 
\textbf{`still\_failing`} --- Explain why the patch was insufficient. What should the next iteration try instead?
\ \ \ \ ```bash
\ \ \ \ evo-dag update-reflection \textbackslash
\ \ \ \ \ \ --node \{candidate\_idx\} \textbackslash
\ \ \ \ \ \ --scenario "<id>" --status "still\_failing" \textbackslash
\ \ \ \ \ \ --root-cause "Agent uses absolute dates instead of relative dates in messages." \textbackslash
\ \ \ \ \ \ --explanation "Prompt rule 'use relative dates' was too weak --- agent still used 'Oct 22'. Partial progress: bullet formatting was fixed." \textbackslash
\ \ \ \ \ \ --prevention-or-next "Stronger rule needed. Consider PreToolUse hook on send\_message for just-in-time reminder."
\ \ \ \ ```
 
\textbf{`still\_passing`} --- For scenarios that use the same tools or components you modified, explain which specific patch change could have caused interference and why it didn't. This is valuable signal for understanding patch safety.
\ \ \ \ ```bash
\ \ \ \ evo-dag update-reflection \textbackslash
\ \ \ \ \ \ --node \{candidate\_idx\} \textbackslash
\ \ \ \ \ \ --scenario "<id>" --status "still\_passing" \textbackslash
\ \ \ \ \ \ --explanation "Uses send\_message but passes because it sends user-provided verbatim content. Our rule only affects agent-composed messages."
\ \ \ \ ```
For scenarios unrelated to the patch, use the batch command:
\ \ \ \ ```bash
\ \ \ \ evo-dag update-reflection --node \{candidate\_idx\} --batch-still-passing "id1,id2,..."
\ \ \ \ ```
 
\{generalization\_workflow\_step\}
 
\#\# Causal Attribution Checklist
 
Before finishing, verify each regressed AND fixed scenario's reflection:
 
1. \textbf{Classified}: Explicitly labeled as TRUE (REGRESSION/FIX) or STOCHASTIC
2. \textbf{Evidence-based}: Cites the specific divergence point in traces
3. \textbf{Diff-linked}: States whether the diff touches the relevant code path
4. \textbf{Actionable}: True regressions have prevention guidance; true fixes explain the causal chain; stochastic cases explicitly state the evidence and implications
 
\#\# Reflection Quality Checklist
 
Before finishing, verify each reflection meets these criteria:
 
1. \textbf{Specific}: References exact tool calls, steps, or code paths
2. \textbf{Causal}: Explains WHY the outcome occurred, not just WHAT happened
3. \textbf{Actionable}: `--prevention-or-next` gives clear guidance for future iterations --- not generic advice like "try harder"
4. \textbf{Distinct}: Each reflection adds unique information (no copy-paste templates across scenarios)
 
\#\# Constraints
 
- Record a reflection for \textbf{every} scenario in the mini-batch.
- Use `evo-dag update-reflection` --- do NOT write reflections to files.
- Do NOT modify source code during reflection --- Session 2 is analysis only.
- Do NOT read raw traces from development set output directories --- only aggregate dev accuracy is available for generalization analysis.
- Do NOT read raw traces from development set dirs (`\_val\_` or `seed\_val\_` output dirs) --- only aggregate accuracy is available.
 
\textbf{CRITICAL}: You MUST run `evo-dag update-reflection` for \textbf{every} scenario in the mini-batch before finishing this session. Use individual calls for fixed/regressed/still\_failing scenarios and `--batch-still-passing` for unaffected passing scenarios. Missing reflections degrade the quality of accumulated lessons for future iterations.
}
\end{tcolorbox}
 
\captionof{figure}{Reflection Session Prompt: agent analyzes per-scenario initial vs.\ re-evaluation outcomes (\texttt{fixed} / \texttt{regressed} / \texttt{still\_failing} / \texttt{still\_passing}), distinguishes true causal effects of the patch from stochastic LLM non-determinism, and records structured reflections that feed back into \texttt{evo-dag show history} for future iterations.}
\label{fig:reflection-session-prompt}

\begin{tcolorbox}[
    enhanced,
    breakable,                       
    colback=white,
    colframe=gray!50,
    boxrule=0.4pt,
    arc=4pt,
    left=8pt, right=8pt, top=4pt, bottom=4pt,
    title=\textbf{Evolution Session Prompt},
    coltitle=white,
    colbacktitle=black,
    fonttitle=\small\bfseries,
    halign title=center,
    attach boxed title to top={yshift=-0.5mm, xshift=-0.4pt},
    boxed title style={
        colback=black,
        colframe=black,
        boxrule=0pt,
        arc=4pt,
        sharp corners=south,
    },
    width=\linewidth,
    %
    pad at break*=2mm,               
    overlay middle and last={
        \node[anchor=south east, font=\footnotesize\itshape, gray]
        at ([xshift=-6pt,yshift=2pt]frame.north east)
        {(continued)};
    },
]
{\fontsize{8pt}{9.5pt}\ttfamily%
\setlength{\parindent}{0pt}%
\setlength{\parskip}{2pt}%
\raggedright%
\obeylines%
\#\# Mandatory Skills
 
You MUST read and follow the SKILL.md for each skill listed below. Do NOT skip or summarize any skill --- execute the full procedure described in each one. These skills are installed at `.claude/skills/<n>/SKILL.md` in the current worktree.
 
| Order | Skill | When | Why |
|-------|-------|------|-----|
| 1 | `history-analysis` | \textbf{Before any other work} (Step 1) | Structured analysis of the full evolution history --- classifies every patch as revert vs. preserve to inform candidate selection |
| 2 | `patch-verification` | \textbf{After any codebase change} (Step 5) | Verifies the prepared base codebase has no runtime errors --- a broken base causes every subsequent session to fail |
 
> \textbf{Enforcement}: If you make any change to the worktree (rsync, cherry-pick, revert, or code edit), you MUST run the `patch-verification` skill's full procedure before finishing. Skipping verification is a critical failure.
 
\#\# Goal
 
Prepare the best possible base codebase for this iteration's patch. The optimization objective is to \textbf{maximize performance across the task distribution}, with development set (dev) accuracy as the proxy.
 
Your selection must be \textbf{data-driven}, based on dev scores, regression analysis, and cross-candidate code comparison --- not on attachment to accumulated patches.
 
\textbf{Anti--Sunk-Cost Principle}: The number of iterations invested in a lineage is NOT a reason to continue it. A long lineage with declining dev scores is a signal to switch, not to persist. Evaluate each candidate by its \textbf{measured performance} and the \textbf{quality of its code}, not by how many patches led to it.
 
\#\# Context
 
- \textbf{Iteration}: \{iteration\}
- \textbf{Phase}: \{phase\}
- \textbf{Current worktree}: `\{worktree\_path\}` (fork of C\{base\_parent\_idx\})
- \textbf{Base parent}: C\{base\_parent\_idx\} (worktree: `\{parent\_worktree\}`)
 
\#\# Candidate Performance (sorted by dev score)
 
\{candidate\_table\}
 
\textbf{Note}: Dev scores are only available for candidates that passed mini-batch acceptance. "pending" means the candidate was not evaluated on the dev set. Use patch history, raw traces, and codebase inspection to assess these.
 
\#\# DAG Topology
 
\{dag\_topology\}
 
\#\# Selection Strategy
 
Choose the base candidate by weighing dev scores, regression history, score trajectory, and code quality together. Dev scores are noisy --- small differences may be evaluation variance, so corroborate with regression counts and code inspection. Do not stay on a lineage just because many patches were accumulated; patches only have value if performance improved.
 
\textbf{Options}:
- \textbf{Continue from latest}: When the lineage is near its peak and stable. Revert any regressions and cherry-pick fixes from other candidates.
- \textbf{Switch parent}: When dev score has dropped well below the best candidate and the trajectory is not recovering. Cherry-pick verified good patches from the old lineage.
- \textbf{Combine candidates}: When different candidates have complementary strengths in non-overlapping areas.
 
\textbf{Key rules}:
- Regressions identified here must be reverted in THIS session. Do NOT defer to Session 1 --- Session 1 only addresses the current mini-batch and will not fix prior regressions.
- When cherry-picking, prefer diff-level precision over copying entire files. Use `evo-dag show edge` to see exact diffs and apply only the relevant changes. Whole-file copies via rsync bring unintended changes.
- Compare top candidates' codebases directly (`diff -rq`) to find cherry-pick opportunities that aren't visible from scores alone.
 
\#\# Workflow
 
\#\#\# 1. Analyze history
 
\textbf{MANDATORY first step.} Run the `history-analysis` skill to build a structured understanding of the full evolution history. Do NOT skip this step or truncate `evo-dag show history` with `head`/`tail`. The skill provides the methodology for reading the complete history without truncation.
 
The purpose of this analysis is to \textbf{classify every patch} in the history into two categories that directly inform candidate selection:
 
\#\#\#\# Patches to revert (caused severe regressions)
- Which patches introduced regressions? Identify the exact iteration, candidate, and code diff that caused each regression.
- How severe is each regression? (How many scenarios regressed? Did dev score drop?)
- Is the regression still present in the current lineage, or was it already fixed by a later patch?
- What files were modified --- so you know exactly what to revert or undo.
 
\#\#\#\# Patches to preserve (drove performance gains)
- Which patches produced the largest improvements in dev score?
- Which patches fixed scenarios that stayed fixed in subsequent iterations (durable fixes vs. fragile ones)?
- Which patches generalized well beyond their target mini-batch?
- What files and patterns were involved --- so you know what to protect when combining candidates.
 
This classification directly drives your selection decision:
- \textbf{Continue from latest}: Revert the harmful diffs while keeping the beneficial ones.
- \textbf{Switch parent}: If too many regressions have accumulated, switch to the best candidate and cherry-pick the preserved patches.
- \textbf{Combine candidates}: If different candidates have complementary preserved patches, combine them.
 
\#\#\# 2. Investigate candidates
 
Use the following sources to decide which candidate(s) to build on:
 
1. \textbf{History analysis output}: The structured summary from Step 1 --- lineage trajectory, lessons, and cross-candidate strengths.
2. \textbf{Specific candidate}: `evo-dag show node <idx>` --- scores, patch intent, verdict, output directories.
3. \textbf{Raw traces}: Read evaluation output and trace files in the iteration output directories listed by `evo-dag show node` or `evo-dag show current-batch`. See CLAUDE.md's \textbf{Iteration Output Structure} section for the file layout.
4. \textbf{Candidate codebases}: Read source files directly in other candidates' worktrees to understand what changed and whether a fix is worth porting. Worktree paths are shown in the candidate table and `evo-dag show node`.
5. \textbf{Cross-candidate code comparison (MANDATORY for top candidates)}: For the top 2-3 candidates by dev score, directly compare their codebases to identify divergent files and cherry-pick opportunities:
\ \ \ \ ```bash
\ \ \ \ diff -rq \{session\_root\}/worktrees/<candidate\_A\_dir>/ \{session\_root\}/worktrees/<candidate\_B\_dir>/ | grep -v '.git'
\ \ \ \ ```
\ \ \ Then read the divergent files in both worktrees to understand which candidate's version is better for each file. This comparison reveals cherry-pick opportunities that are invisible from score data alone.
 
\#\#\# 3. Revert regressions (MANDATORY if continuing from latest)
 
If you chose to continue from the latest candidate, you MUST revert all identified regressions in THIS session. Do NOT defer regression fixes to Session 1 --- Session 1 only addresses the current mini-batch and will not fix prior regressions.
 
For each regression identified in Step 1:
1. \textbf{Locate the exact diff} that caused the regression using `evo-dag show edge <parent> <child>` for the iteration that introduced it.
2. \textbf{Compare with pre-regression code}: Read the same file in the pre-regression candidate's worktree to see what the code looked like before the harmful change.
3. \textbf{Revert only the harmful lines}: Do not revert the entire patch if it also contains beneficial changes. Surgically revert only the lines that caused the regression.
4. \textbf{Preserve beneficial changes}: If a patch contains both good and bad changes, keep the good parts. Check whether specific scenarios were fixed by specific lines in the diff to determine what to preserve.
 
After reverting, run the `patch-verification` skill to ensure no runtime errors were introduced.
 
\#\#\# 4. Prepare the base codebase
 
Each candidate has a \textbf{base parent} (the candidate it was forked from) and optionally \textbf{cherry-pick parents} (other candidates whose code was referenced, copied, or combined). The base parent is set automatically by the outer loop. Cherry-pick parents are what you record here.
 
Your worktree starts as a fork of C\{base\_parent\_idx\} (base parent).
 
\#\#\#\# Switching base entirely
 
If you decided to switch base, copy the new base's entire codebase:
\ \ \ \ ```bash
\ \ \ \ rsync -a --exclude='.git' \{session\_root\}/worktrees/<candidate\_dir>/ ./
\ \ \ \ ```
Then cherry-pick good patches from the old lineage using the diff-level method below.
 
\#\#\#\# Diff-level cherry-pick (preferred over file-level copy)
 
When porting specific fixes from another candidate, use \textbf{diff-level precision} rather than copying entire files. Copying entire files brings unintended changes from the source candidate.
 
1. \textbf{Identify the exact diff}: Use `evo-dag show edge <parent> <child>` to see the exact code changes that constituted the fix you want to port.
2. \textbf{Compare files}: Read the specific file in both the source candidate's worktree and your current worktree. Use `diff` to see differences:
\ \ \ \ ```bash
\ \ \ \ diff \{session\_root\}/worktrees/<source\_candidate\_dir>/<file> ./<file>
\ \ \ \ ```
3. \textbf{Apply only the relevant changes}: Manually apply only the lines from the diff that correspond to the fix. Do not copy unrelated changes that the source candidate may have accumulated.
4. \textbf{Verify}: After each cherry-pick, run the `patch-verification` skill.
 
\#\#\# 5. Verify changes
 
If you made any changes to the worktree (rsync, cherry-pick, revert, etc.), you MUST run the `patch-verification` skill to ensure the codebase has no runtime errors. A broken base codebase will cause every subsequent session to fail --- verification is mandatory.
 
\#\#\# 6. Record selection
 
After preparing the base code, record your selection with `evo-dag update-selection`. List \textbf{all} candidates you referenced --- whether you switched base entirely, cherry-picked specific files, or just used their code as reference for a re-implementation:
\ \ \ \ ```bash
\ \ \ \ evo-dag update-selection \textbackslash
\ \ \ \ \ \ --parent-candidates "<idx1>,<idx2>,..." \textbackslash
\ \ \ \ \ \ --reasoning "<what you took from each and why>"
\ \ \ \ ```
 
\textbf{CRITICAL}: You MUST run `evo-dag update-selection` before finishing this session. Failure to do so will result in lost selection history that future iterations need to understand the DAG lineage.
 
\#\# What Happens Next
 
After this session, the outer loop runs \textbf{Session 1 (Diagnose + Patch)}: the agent receives the current mini-batch's failing scenarios with execution traces, diagnoses root causes, and applies targeted code patches to the worktree you prepared here.
}
\end{tcolorbox}
 
\captionof{figure}{Evolution Session Prompt: data-driven base codebase selection from prior candidates, including history analysis, regression reversion, and diff-level cherry-picking.}
\label{fig:evolution-session-prompt}

\section{Data Licenses}
\label{app:licenses}

The licenses for the datasets used in this paper are as follows:
\begin{itemize}
    \item GAIA2: cc-by-4.0
    \item SWE-Bench Pro: MIT license
    \item Terminal-Bench 2.0: Apache-2.0 license
\end{itemize}
All datasets were used strictly for research purposes and were not utilized in any non-research contexts, particularly for commercial applications.

\section{Limitations and Future Work}
\label{app:limitations}

\proposed{} currently formulates harness optimization as a supervised learning problem, assuming access to a training set of tasks paired with gold answers and a task-level success metric (e.g., pass/fail) that provides a clear signal for each rollout. While this assumption aligns with standard agent benchmarks, real-world deployment settings may lack such ground-truth outcomes, and obtaining them at scale can be prohibitively expensive. A natural extension is therefore to relax this supervision requirement and study unsupervised or weakly supervised harness optimization, in which failure signals are derived from intrinsic cues such as trace-level inconsistencies, tool-call errors, or self-consistency across rollouts rather than external labels. Another direction is to synthesize training data by bootstrapping from agent environments, such as each user’s \texttt{Universe} in GAIA2. Finally, our current scope is limited to stateless, independent tasks; extending \proposed{} to stateful settings, integrating memory and skill curation, and validating the approach across a broader set of LLM families remain important directions for future work.

In terms of real-world deployment, we want to stress the importance of strong governance and human-in-the-loop in the early development stage, namely, automated harness modifications should undergo human review or additional security validation before deployment to production systems.


\end{document}